# Optimal Characteristics of Inspection Vehicle for Drive-by Bridge Inspection

A. Calderon Hurtado[a], E. Atroshchenko[a], K.C. Chang[b], C.W. Kim[b], M. Makki Alamdari[a,*]

[a]*Center for Infrastructure Engineering and Safety, School of Civil and Environmental Engineering, University of New South Wales, Sydney, NSW 2052, Australia*
[b]*Department of Civil & Earth Resources Engineering, Graduate School of Engineering, Kyoto University, Kyoto 615-8540, Japan*

**Abstract**

Drive-by inspection for bridge health monitoring has gained increasing attention over the past decade. This method involves analysing the coupled vehiclebridge response, recorded by an instrumented inspection vehicle, to assess structural integrity and detect damage. However, the vehicles mechanical and dynamic properties significantly influence detection performance, limiting the effectiveness of the approach. This study presents a framework for optimising the inspection vehicle to enhance damage sensitivity. An unsupervised deep learning methodbased on adversarial autoencoders (AAE)is used to reconstruct the frequency-domain representation of acceleration responses. The mass and stiffness of the tyresuspension system of a two-axle vehicle are optimised by minimising the Wasserstein distance between damage index distributions for healthy and damaged bridge states. A Kriging meta-model is employed to approximate this objective function efficiently and identify optimal vehicle configurations in both dimensional and non-dimensional parameter spaces. Results show that vehicles with frequency ratios between 0.3 and 0.7 relative to the bridges first natural frequency are most effective, while those near resonance perform poorly. Lighter vehicles require lower natural frequencies for optimal detection. This is the first study to rigorously optimise the sensing platform for drive-by sensing and to propose a purpose-built inspection vehicle.

## 1. Introduction

Bridges, critical infrastructure in todays society, are significant in maintaining connectivity and supporting every economic activity [1]. However, over time, bridges are exposed to a variety of factors that can degrade their structural integrity, including changes in loading conditions, environmental effects, material fatigue, corrosion, and faulty construction [2]. As a result of these weaknesses, there is increasing concern around the world about maintaining the long-term health and safety of bridges [3, 4].

*Corresponding author
  *Email address:* m.makkialamdari@unsw.edu.au (M. Makki Alamdari)

Systems for structural health monitoring, or SHM, have become an essential instrument for evaluating the state of bridges. These systems are often classified into two primary categories: direct and indirect. Direct SHM systems entail putting a network of sensors directly on the bridge structure to monitor its response continually. Although effective, this method is often costly, complicated, and requires substantial maintenance efforts, making it challenging to implement widely [5, 6, 7]. Conversely, indirect drive-by bridge structural health monitoring systems, which involve using sensors on passing vehicles to assess bridge conditions, offer a more economical alternative by eliminating the need for extensive sensor installations on the structure itself [6, 8, 9].

SHM techniques are distinguished not just by the implementation methods but also by their underlying approach: either physics-based or data-driven. Physics-based methodologies focus on identifying changes in the bridge's physical characteristics, including strain, displacement, natural frequencies, and mode shapes, to detect potential deterioration. These methods are typically rooted in structural dynamics and often require detailed knowledge of the bridge's design and materials [10, 11, 12]. On the other hand, data-driven approaches, as the one used in this work, rely on analysing response data from the bridge without requiring extensive prior knowledge of its structure. These methods leverage statistical and machine learning techniques to process the data and identify patterns indicative of damage [2, 6, 8, 13].

*1.1. Data-driven Drive-by Bridge Monitoring*

Selecting damage-indicative attributes or features from the vehicle's response data constitutes the basis of data-driven methodologies in drive-by bridge monitoring [14]. The presence and severity of damage in the bridge can then be inferred by analysing the extracted features. For instance, processing the vehicle's acceleration data and identifying indicators of structural deterioration has been done using wavelet analysis in combination with pattern recognition algorithms [15]. Similarly, characteristics from acceleration signals have been extracted using Principal Component Analysis (PCA) and input into predictive models for damage identification and severity classification [16, 17].

There is an increasing interest in utilising machine learning and deep learning techniques in the field of indirect structural health monitoring. Techniques including deep autoencoders, semi-supervised learning, and convolutional neural networks (CNNs) have demonstrated significant efficacy in detecting bridge deterioration through the analysis of vehicle-bridge interaction (VBI) data [18, 19, 20, 21]. For instance, adversarial autoencoders have been trained on vehicle acceleration spectrograms to detect damage of different severities in bridges [22], while other approaches have employed convolutional autoencoders to analyse time-frequency spectrums from the vehicle acceleration to identify flexural crack up to 5% of crack depth to beam height ratio [23].

The efficacy of data-driven methodologies is reliant on the sensitivity of the chosen features and the quality of the data [24]. Noise from the road surface and variations in vehicle speed can obscure the underlying bridge-related information, making feature extraction and pattern recognition more complex [25]. Despite these challenges, advances in computational techniques and access to larger datasets have made data-driven methods increasingly viable for real-time bridge health monitoring.

*1.2. Inspection Vehicle in Drive-by Bridge Inspection*

A variety of studies have attempted to determine the characteristics of inspection vehicles (including vehicle model, suspension and tire stiffness and damping, and vehicle mass) to enhance the identification of bridge features to assess bridge damage based on sensor data collected from the vehicle. Kong et al. [26] established an approach for identifying bridge modal parameters through the reaction of a sensing trailer towed by a vehicle. The study extracted natural frequencies and mode shapes using Fast Fourier Transform (FFT) and Short-Time Fourier Transform (STFT). The researchers examined the impact of the trailer's natural frequency, determining that it must be less than the bridge's natural frequency. The researchers determined that reduced trailer speed and a separation of under 2.5 meters between trailers yielded superior identification outcomes. Li et al. [27] applied stochastic subspace identification to extract bridge modal properties from vehicle responses. They found that lower speeds (i.e., 2 $m/s$) improved accuracy, that performance deteriorated when the vehicles natural frequency approached the bridges, and that lighter vehicles (i.e., 100 $kg$) with moderate suspension stiffness were favourable. These results underscore the importance of vehiclebridge frequency ratio and suspension properties in bridge modal identification.

Yang et al. [28] examined the determination of bridge frequencies by the frequency analysis of responses captured by an inspection vehicle. It was discovered that when the vehicle-bridge frequency ratio nears resonance (i.e., $\approx 1$), the determined natural frequency diverges by 60% from the true value. Moreover, they discovered that when the vehicle mass is considerably less than the bridge mass, the fluctuation in the determined bridge frequency is minimal. In a separate investigation, Yang et al. [29] empirically determined bridge frequencies utilising a test vehicle. It was suggested that, to enhance the identification of bridge parameters, the vehicle should possess elevated suspension and tire stiffness, along with rigid tires, to ensure a distinct vehicle-bridge frequency ratio. Shi and Uddin [30] retrieved bridge frequencies using an inspection vehicle and determined that high vehicle damping, a vehicle frequency significantly distant from the bridge frequency, and low vehicle velocity facilitate accurate identification of bridge dynamical characteristics.

Bu et al. [31] proposed a damage assessment approach utilising sensitivity analysis of the dynamic response recorded by a traversing vehicle. The researchers assessed the influence of different VBI system parameters on damage evaluation, determining that a vehicle-bridge mass ratio of 0.015 and a vehicle-bridge frequency ratio of 2.5 are ideal for damage identification. Separately, McGetrick and Kim [32] established a damage assessment methodology employing Morlet wavelet analysis of the data captured by an inspection vehicle. They conducted a parametric analysis to evaluate the impact of vehicle characteristics, including mass and speed, indicating that a vehicle frequency that differs from the bridge frequency, along with a reduced vehicle speed, is advantageous. Zhu et al. [33] proposed a drive-by bridge inspection methodology for damage detection, utilising Newton's iterative approach to refine a damage parameter. They confirmed that vehicle speed and mass influence damage identification accuracy, suggesting that a slower, heavier vehicle enhances this accuracy.

In summary, several authors recommend using inspection vehicles with natural frequencies that differ from those of the target bridge [34]. Additionally, heavy vehicles that induce greater excitation in the bridge, combined with low vehicle velocities during sensing, are suggested to enhance damage assessment accuracy [8]. Although multiple studies have performed parametric studies with limited vehicle properties, aiming to identify the ideal characteristics of inspection vehicles for maximising performance in bridge identification and damage assessment, the optimisation of the inspection vehicle characteristics has not yet been addressed in the literature. This gap is largely due to the stochastic nature of the vehicle-bridge interaction (VBI) problem, influenced by measurement noise and road roughness profiles [31, 35, 36, 37]. To address this, the present study introduces a rigorous inspection vehicle design optimisation framework aimed at improving damage identification based on an unsupervised learning methodology by taking into account the non-deterministic nature of the problem.

*1.3. Research Contribution*

Prior studies on the design optimisation of inspection vehicles for bridge damage assessment using drive-by inspection have been limited. The optimisation problem has not been sufficiently addressed, as the majority primarily concentrate on conducting parametric studies with restricted vehicle configurations. This research seeks to address this deficiency by offering a rigorous optimisation methodology for the inspection vehicle, focusing on non-dimensional metrics associated with both the vehicle and bridge characteristics. Furthermore, pragmatic suggestions for the design of the inspection vehicle are provided. These guidelines, applicable to beam-type bridges with varying dynamic characteristics, serve as practical instruments for enhancing bridge health monitoring through drive-by inspections in real-world contexts. This design optimisation framework relies on the damage assessment efficacy of an unsupervised deep-learning approach founded on adversarial autoencoders [22].

The subsequent sections of the paper are organised as follows: Section 2 describes the numerical model for vehicle-bridge interaction. Section 3 establishes the numerical case study. Section 4 outlines the damage assessment technique and the proposed optimisation framework for inspection vehicles. The findings, covering the design recommendations and validation of the inspection vehicle, are provided in Section 5. Section 6 presents a benchmark study using the purpose-built sensing vehicle on a healthy structure, while Section 7 outlines directions for future work. Finally, Section 8 concludes the manuscript with a summary of findings and suggestions for further research.

## 2. Vehicle-Bridge Interaction (VBI) Model

This section presents the dynamic equations governing the vehiclebridge interaction, which are used to generate data for the numerical case study discussed in Section 3.

*2.1. Overview of VBI model*

This study uses a half-car model as presented by [22] in the numerical simulations, as depicted in Figure 1.

The half-car model incorporates two degrees of freedom: vertical motion and pitch rotation, represented by $z_v(t)$ and $\theta_v(t)$, respectively. These variables are used to determine the chassis displacement at each axle position, denoted $z_{a_i}$. The vehicle is characterised by eight parameters: the stiffness $k_{v_i}$ and damping $c_{v_i}$ of the combined tyre-suspension system at the $i$-th axle, the total mass $m_v$, the moment of inertia $I_v$, and the distances to the axles $d_i$. The model assumes a constant vehicle speed $v$ and perfect contact between the tyres and the road surface. Chassis-mounted accelerometers measure the acceleration responses $\ddot{z}_{a_i}$.

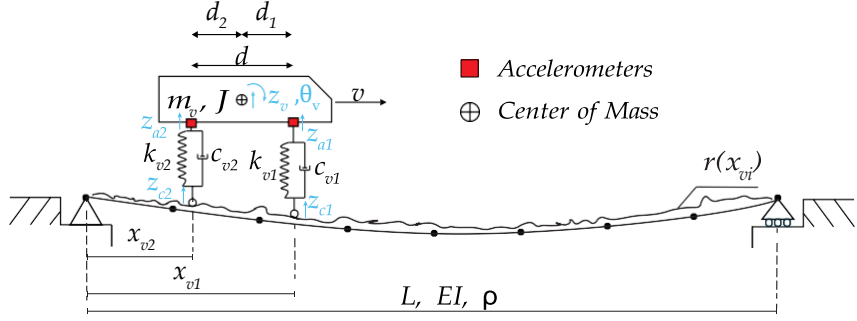

Figure 1: Representation of the vehicle-bridge interaction model employed in the numerical simulation.

The bridge is modelled separately as a simply supported Euler-Bernoulli beam, featuring a stochastic road profile $r(x_{v_i})$ at the axle positions [22]. Additionally, the analysis considers a single inspection vehicle crossing the bridge in the absence of other traffic. Further information on the finite element model is available in Appendix Appendix A.

*2.2. Contact Point Response*

Previous research has demonstrated the benefits of using the contact point (CP) response of the axles instead of the direct axle force during bridge drive-by inspections [38, 39, 40]. The main advantage of relying on the CP response lies in its ability to reduce the influence of the vehicle on the measured time signal. The CP displacement at the $i$-th axle can be expressed in terms of the bridge response and the road roughness at the contact point as follows [23]:

$$z_{c_i}(t) = Z_b^{x_{v_i}}(t) + r(x_{v_i}) \tag{1}$$

where, $z_{c_i}(t)$ denotes the contact point displacement, $Z_b^{x_{v_i}}(t)$ is the bridge response at the CP location, obtained from the VBI model described in Section 2.1, and $r(x_{v_i})$ is the road profile at that location. However, direct measurement of the vehicle's CP response is not practical in field conditions.

To address this, Hurtado et al. [23] proposed a formulation that estimates the CP response using acceleration data measured at the vehicle axles. The CP displacement at axle $i$ is computed as:

$$z_{c_i}(t) = \int_0^t \frac{e^{-k_{v_i}(t-\tau)/c_{v_i}}}{c_{v_i}} g_i(\tau)\, d\tau + z_{c_{i_o}}(t) e^{-k_{v_i} t/c_{v_i}} \tag{2}$$

where the input function is defined as $g_i(\tau) = m_{v_i}\ddot{z}_{a_i}(t) + c_{v_i}\dot{z}_{a_i}(t) + k_{v_i}z_{a_i}(t)$, and $z_{a_i}(t)$ represents the displacement of the axle. The initial CP displacement $z_{c_{i_o}}(t)$ is assumed to be zero, which is reasonable when the road profile $r(x_{v_i})$ is zero at the bridge supports. The term $m_{v_i}$ is the portion of the vehicle mass assigned to axle $i$, $t$ is the time at which the axle is on the bridge, and $\tau$ is a dummy variable used in the convolution process when computing the inverse Laplace transform. This formulation was presented and validated against numerical drift and noise in [23], where the authors added 5% white noise to the acceleration signal. Figure 2 shows a comparison of axle displacement response $z_{a_i}$ obtained via the proposed formulation and the response obtained from the VBI FEA.

The stochastic road roughness profile, $r(x)$, is defined as type A road roughness in accordance with ISO8606 [41]. Information regarding the formulation of the road roughness profile is available in [23]. Figure 3 illustrates a typical example of a type A road profile utilised in this study.

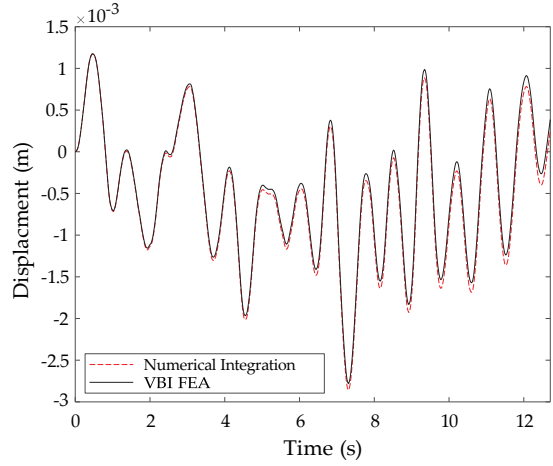

Figure 2: Comparison between ground-truth front axle displacement response and the numerical displacement response obtained with the proposed formulation.

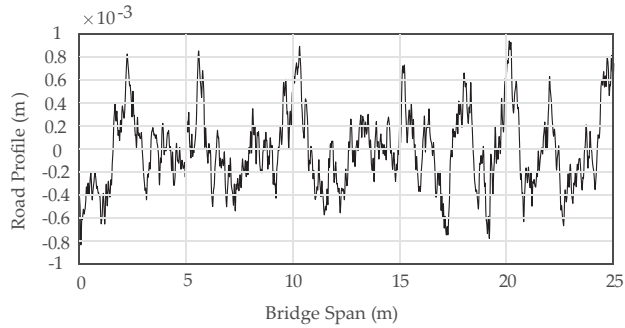

Figure 3: Illustration of class A stochastic road surface profile considered for the beam model.

To represent structural damage in the simulations, this study adopts the simplified cracked beam model proposed by [42]. In this model, damage is introduced by applying a linear reduction in stiffness over a length of $1.5h$ from the crack location, where $h$ denotes the height of the beam. At the point of damage, the moment of inertia is given by $I_c = b(h - h_c)^3/12$, where $b$ is the beam width and $h_c$ is the depth of the crack. Additional information on this modelling approach can be found in [42].

## 3. VBI Case Study

This section presents a description of the numerical case study for which the optimisation of the inspection vehicle properties is conducted in Sections 4 and 5. The bridge designated for inspection is introduced, together with the design specifications of the vehicles employed for the inspection process.

### 3.1. Bridge Model

The bridge model is founded on the research conducted in [22, 23]. The properties are shown in Table 1.

Table 1: Physical properties of the bridge to be inspected.

| Property | Value | Property | Value |
|---|---|---|---|
| Span length, $L$ | 25 $m$ | Cross section area, $A$ | 16.68 $m^2$ |
| Young's modulus, $E$ | $3.5 \times 10^{10}$ $N/m^2$ | Second moment of inertia, $I_o$ | 1.39 $m^4$ |
| Mass per unit length, $\mu_b$ | 18,358 $kg/m$ | Damping ratio, $\xi$ | 3% |
| First three bridge's natural frequencies, $f_b$ | 4.09 $Hz$, 16.36 $Hz$ and 36.8 $Hz$ | | |

Two distinct bridge states are examined: (1) a healthy condition *HN*, and (2) a damaged condition *DM*, characterised by a crack depth to beam height ratio of $a_c$ = 10% situated at mid-span. The damage is classified as low severity and results in a 1.9% reduction in the first natural frequency of the structure. It is presumed that if the ideal inspection vehicle can detect low-severity damage, it will also be capable of identifying high-severity damage.

### 3.2. Vehicle Model

In this study, the inspection vehicle characteristics are optimised to maximise the damage assessment performance. The mass $m_v$ and stiffness $k_v$ of the linked tyre-suspension system are optimised in Sections 4 and 5. The characteristics of the vehicle design space are delineated in Table 2, where the non-dimensional parameters associated with the vehicle properties to be optimised (i.e., vehicle-bridge mass ratio $\mu$, and vehicle-bridge frequency ratio $\beta$) are also present.

Table 2: A summary of inspection vehicle physical properties design space.

| Property | Value |
|---|---|
| Vehicle mass*, $m_v$ | [50 : 20,000] $kg$ |
| Axle damping, $c_v$ | $1 \times 10^4$ $Ns/m$ |
| Vehicle inertia, $I_v$ | 93,234 $kg \cdot m^2$ |
| Axle stiffness*, $k_v$ | [0.03 : 8] $\times 10^6$ $N/m$ |
| C.G to front axle distance, $d_1$ | 2.375 $m$ |
| C.G to rear axle distance, $d_2$ | 2.375 $m$ |
| Bouncing natural frequency, $f_z$ | [0.32 : 4.95] $Hz$ |
| Vehicle-bridge mass ratio, $\mu = m_v/(\mu_b L)$ | [0.0001 : 0.04] |
| Vehicle-bridge frequency ratio, $\beta = f_z/f_{b_1}$ | [0.07 : 1.2] |

* Properties to be optimised.

The mass and stiffness design space presented in Table 2 are chosen based on the inspection vehicles used in previous works [15, 28, 31, 43], and adjusted to fit a wider range of frequency ratios, which includes $\beta$ = 1 and $\beta$ > 1. Additionally, the random sampling of the inspection vehicles is performed by a *latin hypercube* following a similar procedure to the one presented in [20]. This work aims to study the relation between the damage assessment performance of the inspection vehicle with $m_v$ and $k_v$, and so all the other properties remain unchanged.

The vehicle's centre of mass is considered to be situated at the midpoint between the front and rear axles. Furthermore, the front and rear tyre-suspension systems are regarded as identical. Furthermore, it is anticipated that the vehicle will maintain a consistent velocity during each crossing of the bridge. The speed of each pass is randomly determined by a normal distribution with a mean value of 2 $m/s$ and a standard deviation of 0.2 $m/s$, as demonstrated in [22, 23].

It is widely acknowledged that the effectiveness of drive-by bridge inspection improves when the inspection vehicle travels at a low speed [8]. However, it is crucial to take the practical implications into account. Conducting bridge inspections with a slow-moving vehicle may require full or partial bridge closures, adding a challenge to practical implementation. This aspect is not addressed in this study.

It is also acknowledged that the physical aspects of the car may impact the effectiveness of the damage assessment. This study aims to serve as a proof of concept for optimising the inspection vehicle's design to maximise damage assessment performance, offering recommendations for broader applications in the future.

## 4. Proposed Framework

The methodology for optimising inspection vehicle design that maximises damage assessment performance is presented in this section. First, Sections 4.1 and 4.2 present the data collection and pre-processing, respectively. An unsupervised deep learning methodology based on Adversarial Autoencoders, [22] is used in the damage assessment process and is discussed in Section 4.3. The inspection vehicle design optimisation framework is described in Section 4.4. Figure 4 shows the methodology for optimising the inspection vehicle characteristics.

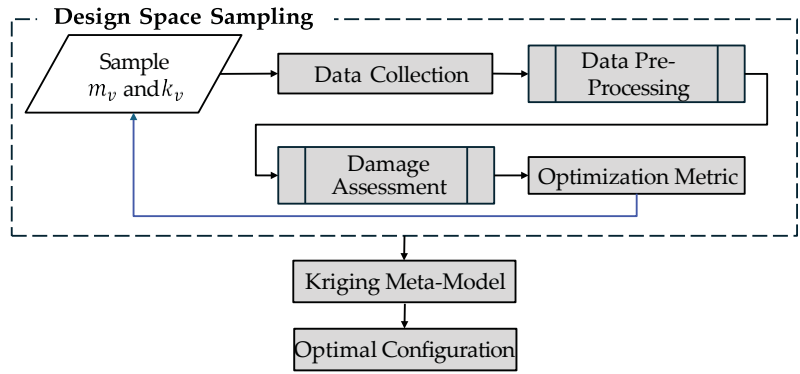

Figure 4: Schematic diagram of sensing vehicle optimisation.

*4.1. Numerical Dataset*

Two bridge conditions (*HN* and *DM*) are considered in this investigation. The response of the inspection vehicle while it travels across the bridge at a constant speed is calculated using the VBI model presented in Section 2. The Newmark-Beta method is employed for time integration, with a time step of 0.001 seconds. The recorded acceleration responses from an inspection vehicle are illustrated in [22].

This analysis suggests that the inspection vehicle routinely traverses the bridge as the bridge's condition deteriorates over time. It also assumes that the dynamic properties of the inspection vehicle remain unaltered throughout multiple passages over the bridge.

To perform the inspection of vehicle characteristics optimisation, 1,500 vehicles are randomly generated. The CP response is collected in accordance with the procedure outlined in Section 2.2, and each inspection vehicle passes over the healthy bridge individually 500 times. Additionally, for each inspection vehicle, 100 CP responses over a damaged bridge are gathered, which corresponds to the same number of healthy samples for testing. In this work, a data-driven unsupervised deep-learning framework (presented in Section 4) is used for damage assessment, which is trained with 80% of the healthy samples. The remaining 20% of the healthy samples are used for testing together with the 100% of the damaged samples (see Section 4.3). It is important to note that each bridge crossing by an inspection vehicle involves a random Type A road roughness profile. This significantly increases the complexity of the problem compared to scenarios where a consistent road roughness profile is applied across all inspection vehicle runs.

*4.2. Data Pre-Processing*

The five primary phases of the signal pre-processing, depicted in Figure 5, are carried out over the data gathered from each inspection vehicle. Each step is described next:

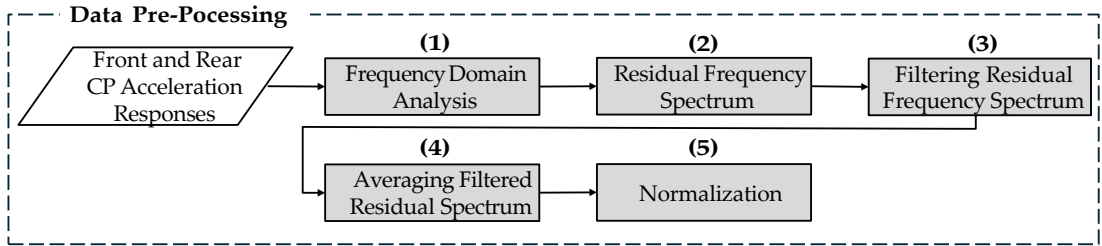

Figure 5: Data pre-processing methodology.

(1) **Frequency domain analysis:** The frequency spectrum of the front and rear CP acceleration responses is obtained through Welch's power spectral density (i.e., $F_{front}(\omega)$ and $F_{rear}(\omega)$, respectively).

(2) **Residual frequency spectrum:** Road roughness effect is reduced by obtaining the residual frequency spectrum $F_{res}(\omega)$, which is defined as $|F_{front}(\omega) - F_{rear}(\omega)|$ [44]. Figure 6 shows an example of the frequency domain representation of the front and rear CP accelerations and its corresponding residual frequency spectrum. It is evident that by doing the subtraction, the bridge's natural frequencies become evident.

(3) **Filtering the residual frequency spectrum:** $F_{res}(\omega)$ is filtered to extract the pertinent frequency components associated with the bridge by choosing a suitable range of the frequency spectrum, e.i., $F'_{res}(\omega) \rightarrow F_{res}(\omega_1 : \omega_2)$. A preliminary numerical model of the bridge under study can easily provide the prior knowledge from the bridge frequency spectrum, which is the basis for selecting this frequency range ($\omega_1 : \omega_2$). In this work, $\omega_1 = 3Hz$ and $\omega_2 = 5Hz$.

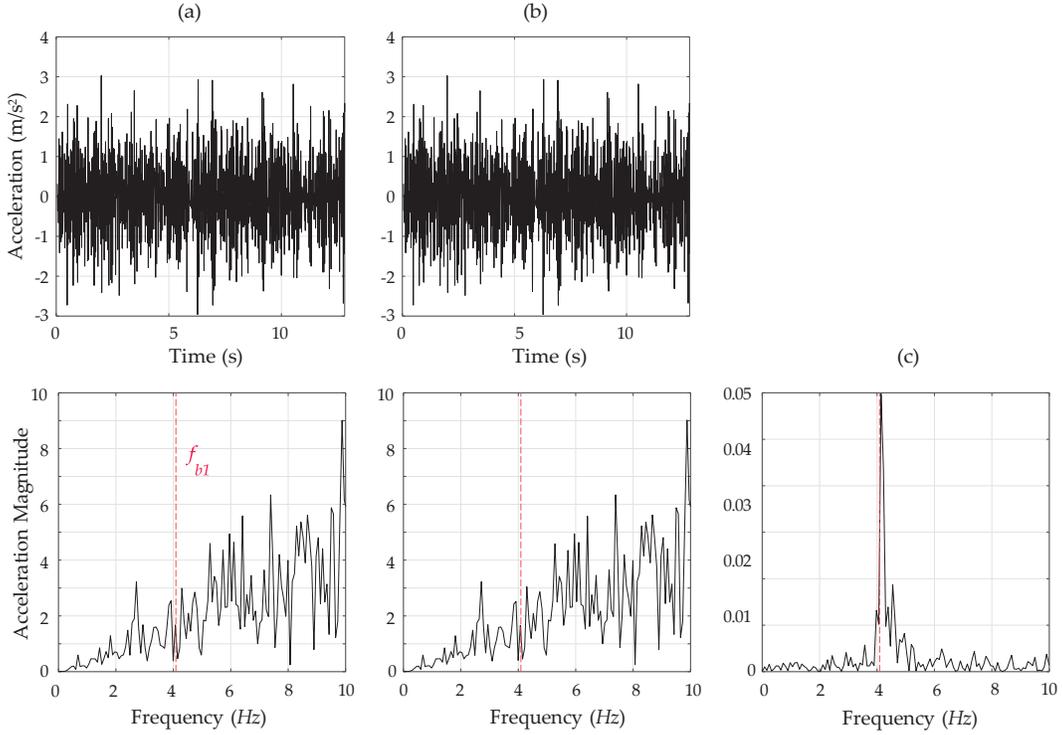

Figure 6: Frequency spectrum of the CP acceleration responses from the inspection vehicle, with $k_v = 4 \times 10^5$ $N/m$ and $m_v = 16,200$ $kg$. (a) Displays the front CP acceleration response and its corresponding frequency spectrum. (b) Displays the rear CP acceleration response and its corresponding frequency spectrum. (c) Displays the residual frequency spectrum.

(4) **Averaging the filtered residual spectrum:** Signal averaging is done as $F_{avg}(\omega) = \frac{1}{k}\sum_{i=1}^{k} F'_{res_i}(\omega)$, where $k$ is the number of vehicle runs from each bridge condition that are taken into account. This is done using $F'_{res}(\omega)$ from numerous runs of the vehicle. As reported in [22], $k = 30$ is used for this investigation. Figure 7 shows an example of the filtered and averaged residual spectrum, in which the variation of the bridge's natural frequency due to damage can be clearly identified.

(5) **Normalization:** Lastly, the averaged data obtained in step (4) for the two bridge conditions (*HN* and *DM*) is normalised with respect to the maximum spectral amplitude of the healthy state dataset (*HN*), scaling the values to the range [0,1], as follows:

$$F(\omega) = \frac{F_{avg}(\omega) - \min(F_{avg_{HN}}(\omega))}{\max(F_{avg_{HN}}(\omega)) - \min(F_{avg_{HN}}(\omega))} \quad (3)$$

where the normalised acceleration spectrum is denoted by $F(\omega)$. The processed data, $F(\omega)$, is provided to the AAE model used for damage assessment (see Section 4.3).

*4.3. Damage Identification Framework*

The damage assessment methodology employed in this study is derived from the unsupervised deep learning methodology, which employs an adversarial autoencoder (AAE) [22]. The authors of this study illustrated the effectiveness of the damage assessment framework in numerical and experimental case studies by contrasting it with state-of-the-art methodologies. The damage assessment framework, as illustrated in Figure 8, is in accordance with the guidelines outlined in [22].

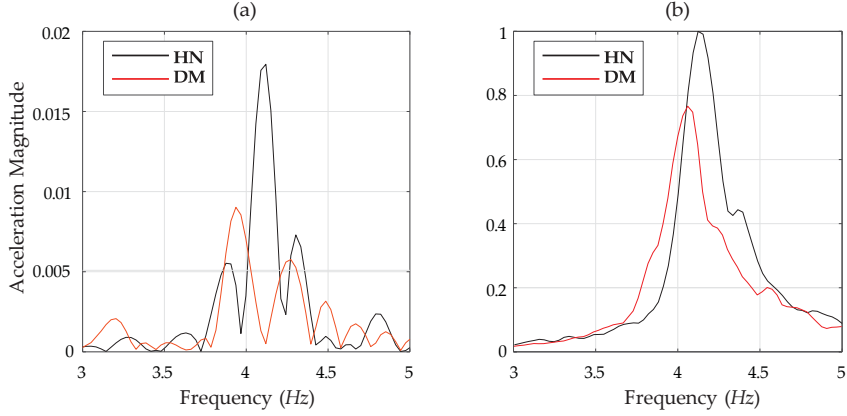

Figure 7: Filtered and averaged frequency spectrum from inspection vehicle with $k_v = 4 \times 10^5$ $N/m$ and $m_v = 16,200$ $kg$. (a) Frequency spectrum without averaging. (b) Frequency spectrum after averaging and normalisation ($k = 30$).

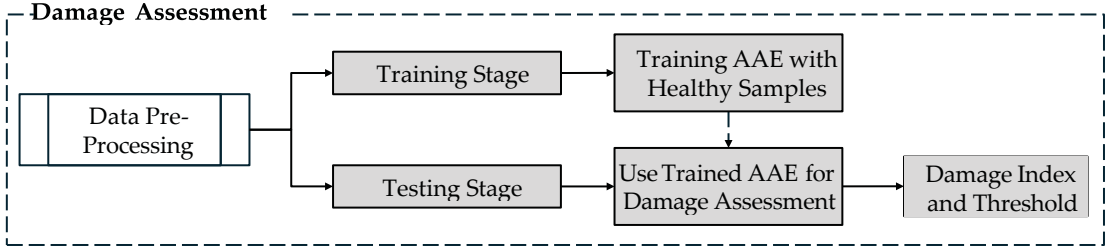

Figure 8: Damage assessment framework.

AAE is a variation of the traditional autoencoder (AE), an unsupervised learning architecture designed to reconstruct its input. An AE consists of two main components: an encoder and a decoder. The encoder reduces the input dimension $n$ of $\mathbf{x} \in \mathfrak{R}^n$ to a lower-dimensional latent representation $\mathbf{y} \in \mathfrak{R}^p$, where $p < n$. The decoder then attempts to reconstruct the original input, producing an output $\bar{\mathbf{x}} \in \mathfrak{R}^n$ [45, 46]. These transformations are defined by the functions $\mathbf{y} = \underline{H}(\mathbf{Wx} + \mathbf{b})$ and $\bar{\mathbf{x}} = G(\underline{\mathbf{W}}\mathbf{y} + \underline{\mathbf{b}})$, where H() and G() are activation functions, and $\mathbf{W}$, $\underline{\mathbf{W}}$, $\mathbf{b}$, and $\underline{\mathbf{b}}$ represent the weights and biases. These parameters are optimised through back-propagation by minimising the reconstruction error between the input and output. Further technical details are provided in [22].

In contrast to a standard AE, the AAE introduces an additional component: a discriminator network that regularises the latent space $\mathbf{y}$. This regularisation aims to align the aggregated posterior distribution $q(\mathbf{y})$ with a chosen prior distribution $p(\mathbf{y})$. The aggregated posterior is defined in Equation 4, where $p_d(\mathbf{x})$ represents the input data distribution [47]. The discriminator estimates the likelihood that a given latent variable $\mathbf{y}$ originates from the prior distribution. Simultaneously, the encoder is trained to fool the discriminator by producing latent representations that mimic the prior. The resulting latent code $\mathbf{y}$ is then used by the decoder to reconstruct the original input [23].

$$q(\mathbf{y}) = \int q(\mathbf{y}|\mathbf{x})p_d(\mathbf{x}) \, d\mathbf{x} \qquad (4)$$

The discriminator network is trained to distinguish between real samples $\bar{\mathbf{y}} \sim p(\mathbf{y})$ drawn from the prior distribution and latent representations $\bar{\mathbf{y}} \sim q(\mathbf{y})$ generated by the encoder [47]. This is achieved by minimising the discriminator loss, $L_D$, which is defined as follows:

$$L_D = \log(D(\bar{\mathbf{y}})) + \log(1 - D(\mathbf{y})) \qquad (5)$$

where the discriminator receives input from $\bar{\mathbf{y}}$ and $\mathbf{y}$. The Sigmoid layer at the end of the discriminator network defines the output of the discriminator, $D(\mathbf{y})$ and $D(\bar{\mathbf{y}})$, within the range of (0, 1). The probability that each sample belongs to the specified prior distribution $p(\mathbf{y})$ is represented by the discriminator output [47]. The generator loss, $L_G$, is determined from Equation 6 after $\mathbf{y}$ is mapped to $\bar{\mathbf{x}}$ through the decoder network. The networks are then updated through back-propagation [23].

$$L_G = (\mathbf{x} - \bar{\mathbf{x}})^2 - \log(D(\mathbf{y})) \qquad (6)$$

The network architecture details are presented in [22], while the general AAE structure is depicted in Figure 9.

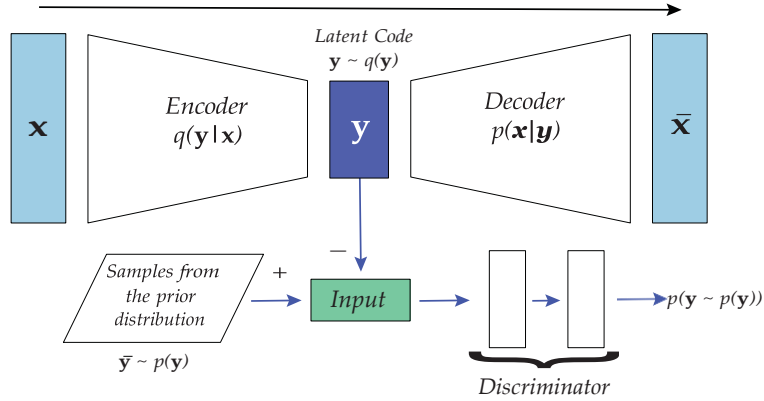

Figure 9: A typical AAE structure [48].

In this work, the input to the AAE $\mathbf{x}$ corresponds to the pre-processed data, $F(\omega)$, generated in Section 4.2. Specifically, 400 preprocessed healthy samples were used to train the AAE model, while testing was conducted using 100 samples for each bridge condition, such as *HN* and *DM*. As previously mentioned, the AAE architecture used in this work is based on the work by Hurtado et al. [22]. The model hyper-parameters were selected to balance convergence stability and computational efficiency. Table 3 summarises the AAE architecture.

*4.3.1. Damage Index*

The damage index (DI) corresponding to each sample is calculated using the Mean Squared Error (MSE) between the original sample $\mathbf{x}_i$ and its reconstruction $\bar{\mathbf{x}}_i$, as shown in

Table 3: Summary of AAE structure.

| Parameter | Value |
| --- | --- |
| Encoder | 256 → 128 → 16 → 8 |
| Decoder | 8 → 16 → 128 → 256 |
| Discriminator | 256 → 128 → 1 |
| Latent code size | 8 |
| Batch size | 16 |
| Epochs | 1,000 |
| Learning rate | $2 \times 10^{-4}$ |

Equation 7, where $i$ denotes the index of the sample.

$$DI = (\bar{\mathbf{x}}_i - \mathbf{x}_i)^2 \tag{7}$$

Furthermore, the $90^{th}$ percentile is used to construct a damage threshold based on the DI distribution of healthy samples (refer to Figure 10). This limit is based on related investigations [49] and helps prevent outliers from being mistakenly categorised as healthy due to factors like measurement noise or variational operation of the bridge.

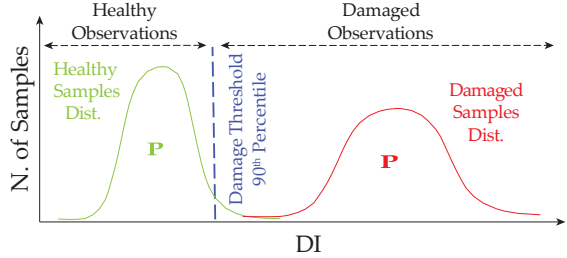

Figure 10: Schematics of the definition of damage threshold based on the $90^{th}$ percentile of the healthy samples DI distribution.

Once the damage threshold is determined, the accuracy based on the classification of the samples as healthy (i.e., *DI <= Threshold*) and damaged (i.e., *DI > Threshold*) is shown in Equation 8

$$\text{Accuracy} = \frac{TP + TN}{TP + TN + FP + FN} \tag{8}$$

where *TP* corresponds to the true-positives, *TN* refers to true-negatives, *FP* are the false-positives, and *FN* correspond to false-negatives. It is worth noting that the classification accuracy depends on the threshold selection. As previously mentioned, the $90^{th}$ percentile selection is based on previous investigations [23, 49].

*4.4. Inspection Vehicle Optimisation*

The inspection vehicle properties are optimised by sampling 1,500 vehicles with different $m_v$ and $k_v$, as presented in Section 3.2, using a *Latin hypercube*. For each of the 1,500 inspection vehicles sampled, the methodology presented in Sections 4.1 and 4.2 is performed.

The Wasserstein distance, $W_d$, between the *HN* and *DM* DI distributions, presented in Section 4.3.1, serves as the metric to be optimised. This metric measures the "work" needed to convert one distribution to another. The first order Wasserstein distance, $W_d(P_1, P_2)$, between two probability distributions, $P_1$ and $P_2$, defined on the metric space P, is expressed as follows [50]:

$$W_d(P_1, P_2) = \inf_{\lambda \in \Lambda(P_1, P_2)} \int ||x_1 - x_2|| d\lambda(x_1, x_2) \quad (9)$$

where $\Lambda(P_1, P_2)$ represents all the joint distributions $\lambda$ with marginals $P_1$ and $P_2$, and $||x_1 - x_2||$ corresponds to the euclidean distance between points $x_1$ and $x_2$, from $P_1$ and $P_2$, respectively, in the metric space P [50]. In the context of this study, $P_1$ denotes the *HN* DI distribution, $P_2$ denotes the DI distributions from the *DM* damage scenario and P is the DI space.

Once the $W_d$ is obtained for each of the 1,500 inspection vehicles sampled, a Kriging meta-model is fitted to the data to provide a simplified approximation between the input (i.e., $m_v$ and $k_v$), and the output (i.e., $W_d$) of the model. Such approximation is called $\overline{W}_d$. The meta-model used in this research is based on the work developed in [51, 52, 53]. Kriging was selected as it is particularly suitable for capturing nonlinear relationships with limited data and provides a reliable global approximation of the objective function. In addition, comparative studies have shown that Kriging generally outperforms other surrogate models, such as polynomial regression or radial basis functions, especially when modelling complex engineering responses with limited datasets, making it especially suitable for the present optimisation task [54, 55]. Finally, the optimal vehicles are identified from this meta-model using Particle Swarm Optimisation (PSO), with the hyper-parameters presented in Table 4.

Table 4: PSO hyper-parameters.

| Hyper-parameter | Value |
| --- | --- |
| Swarm size | 50 |
| Max. Number of iterations | 1000 |
| Inertia Range | (0.5, 0.9) |
| Cognitive Coefficient | 1.49 |
| Social Coefficient | 1.49 |

The optimisation function aims to maximise the distance between the two distributions, which simultaneously signifies a better damage assessment performance. The objective function is defined in Equation 10.

$$\max_{m_v, k_v} \overline{W}_d(P_1(m_v, k_v), P_2(m_v, k_v)) \quad (10)$$

It is essential to note that this study aims to present design guidelines for inspection vehicles. These guidelines are intended for direct application in real-world scenarios, eliminating the need to fully optimise the inspection vehicle for each target bridge. This is possible because the optimisation results are expressed in a non-dimensional space (see Section 5.2) that captures the relationship between vehicle and bridge properties, thereby enabling the generalisation of the guidelines to any VBI system. Additionally,

Furthermore, the scope of this study is to present design guidelines for inspection vehicles targeting small-scale, simply supported, single-span bridges. The applicability of the proposed framework to more complex structural systems remains a subject for future research.

## 5. Results and Discussion

In this section, the results of optimising the inspection vehicle properties are presented. First, Section 5.1 introduces the damage assessment framework for one of the vehicles used in the sampling stage, as shown in Figure 4. Section 5.2 presents the optimisation metric, $\overline{W}_d$, for the inspection vehicle design space described in Section 3.2, where a Kriging meta-model is constructed. Section 5.3 compares the damage assessment performance of various inspection vehicles located in the best-performing and worst-performing regions of the obtained meta-model. Finally, the performance of the optimal inspection vehicle parameters is further validated under a different damage location in Section 5.4.1 and with a different target bridge in Section 5.4.2.

### 5.1. Demonstration of Damage Assessment Framework

In this section, the damage assessment performance using AAE is validated, considering the setup of one inspection vehicle. Specifically, the inspection vehicle used in [22, 23] is investigated. In those studies, the mass, $m_v$, and stiffness, $k_v$, of the inspection vehicle are set to 16, 200 $kg$ and $4 \times 10^5$ $N/m$, respectively, which correspond to a mass ratio $\mu = 0.035$ and a frequency ratio $\beta = 0.27$. Figure 11 shows the damage index distribution from the HN and DM bridge conditions obtained from the damage assessment framework based on AAE.

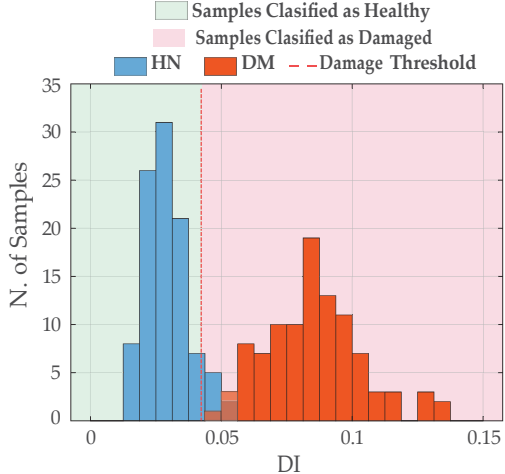

Figure 11: Damage assessment performance of inspection vehicle presented in [22, 23].

It is evident from Figure 11 that the vehicle effectively identifies the damage condition analysed in this study (i.e., $a_c$ = 10% at mid-span). Using a threshold of $DI$ = 0.042, derived from the 90[th] percentile of healthy samples $DI$, the classification accuracy for this particular vehicle configuration is 95%, with an $F_1$ score of 95.2%. Furthermore, the Wasserstein distance $W_d$ between the healthy (HN) and damaged (DM) $DI$ is 0.0559. It is worth noting that the robustness of the damage assessment performance of the vehicle is attributable to the fact that it does not resonate with the bridge, as indicated by $\beta$ = 0.27. Various authors have highlighted this observation [31, 32] and is further discussed in Section 5.2.

## 5.2. Optimisation of Inspection Vehicle

In this section, the optimisation of the inspection vehicle properties (i.e., $m_v$ and $k_v$) is performed. Following the procedure outlined in Section 4.4, the damage assessment performance of 1,500 different inspection vehicles (as detailed in Section 3.2) is evaluated. Then, a Kriging meta-model is constructed to approximate the $W_d$ values for each vehicle (i.e., $\overline{W}_d$). A PSO is subsequently conducted on the developed meta-model to obtain the optimal vehicle characteristics. Figure 12 presents the prediction accuracy of the fitted Kriging meta-model on the test set. The results indicate that the model accurately captures the behaviour of the design space.

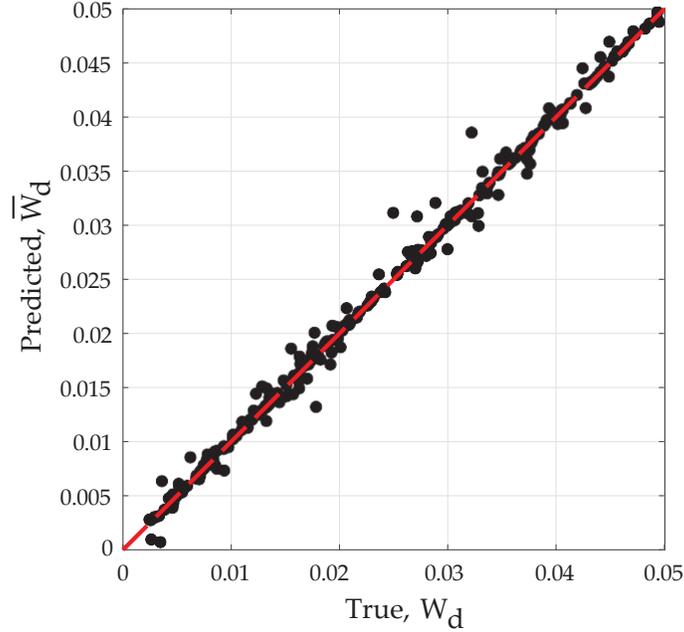

Figure 12: Validation of the Kriging meta-model ($R^2$ = 0.985). The red dashed line indicates the ideal scenario of perfect prediction accuracy. The black dots represent the comparison between the predicted Wasserstein distances ($\overline{W}_d$) and the true values ($W_d$) obtained from the test set.

Figure 13 presents the isocurves representing $\overline{W}_d$ between *HN* and *DM* using the Kriging meta-model, which was trained and tested using an 80:20 ratio from the dataset of 1,500 vehicles. Figure 13a illustrates $\overline{W}_d$ in a dimensional space (i.e., $k_v$ vs. $m_v$), while Figure 13b presents $\overline{W}_d$ in a non-dimensional space, depicting the relationship between $\beta$ and $\mu$ (see Table 2). As shown in Figure 13a, vehicles with lower tyre-suspension stiffness (i.e., $k_v \approx 2 \times 10^6$ *N/m*) and higher vehicle mas (i.e., $m_v \approx 1.8 \times 10^4$ *kg*), exhibit a better damage assessment performance, reflected by larger $\overline{W}_d$ value. This observation, consistent with findings in the literature, is explained by the fact that heavier vehicles induce greater excitation amplitudes in the bridge. Conversely, vehicles with higher $k_v$ and lower $m_v$ have a low damage assessment performance, as indicated by lower $\overline{W}_d$ values.

Similarly, after analysing the damage assessment performance of the inspection vehicles in a non-dimensional space, as shown in Figure 13b, it can be concluded that inspection vehicles whose natural frequency, $f_z$, is close to the first natural frequency of the bridge, $f_{b_1}$ (i.e., $\beta \approx 1$) exhibit the poorest damage assessment performance across the entire design space.

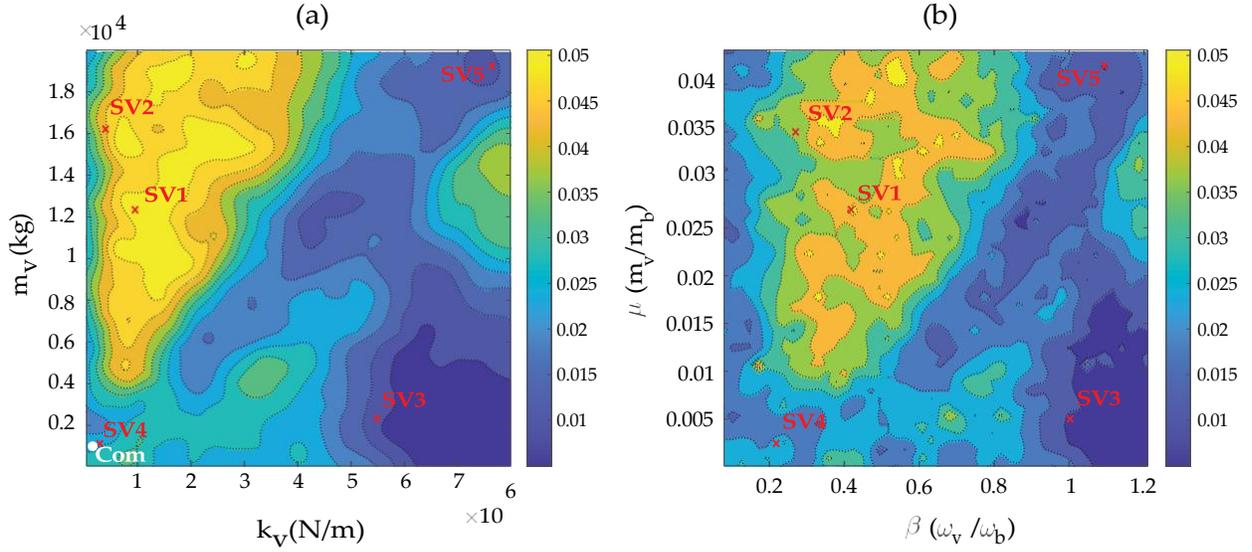

Figure 13: Isocurves representing the predicted $\bar{W}_d$ from the Kriging meta-model between the *HN* and *DM* samples from various inspection vehicle configurations. The red *x*'s represent the sensign vehicles compared in this study. The white *dot*'s represent typical commercial vehicles

Additionally, in the range $0.3 < \beta < 0.7$, there are inspection vehicles that demonstrate exceptional damage assessment performance. However, a constraint on vehicle mass exists. For the lower values of $\beta$ within this range, multiple effective inspection vehicle configurations are viable (i.e., $0.015 < \mu < 0.035$). On the other hand, for larger $\beta$ values within the range, only higher $\mu$ values provide good inspection vehicle performance, meaning large vehicle masses.

Once the meta-model is constructed in the non-dimensional space (see Figure 13b), the optimisation of the inspection vehicles' properties is performed using PSO, following the indications presented in Section 4.4. It is obtained that the optimal vehicle configuration is the inspection vehicle SV1, represented with a red cross in both Figures 13a and 13b. The vehicle properties associated with SV1 are presented in Table 5.

Table 5: Optimal inspection vehicle characteristics.

| **Optimal Inspection Vehicle** | $m_v$ (kg) | $k_v$ (N/m) | $\mu$ | $\beta$ | $\bar{W}_d$ |
|---|---|---|---|---|---|
| SV1 | 12,340 | 9.53 $\times 10^5$ | 0.0269 | 0.418 | 0.0782 |

Furthermore, as the results presented in this study are expressed in a non-dimensional space, they provide valuable guidance and recommendations for designing inspection vehicles applicable to any VBI system used in drive-by bridge inspections. It is important to note that selecting optimal vehicles depends on the damage assessment metric employed. Alternative methodologies may result in different optimal inspection vehicle configurations. Moreover, although many commercial vehicles cannot be modified for bridge inspection purposes, this study is envisioned as a design framework for developing specialised autonomous vehicles for drive-by bridge inspection, an area in which this research group is actively engaged.

Representative parameters of commercially available vehicles have also been considered for comparison. Typical passenger cars exhibit sprung masses in the range of 1,000–1,800 $kg$ with suspension stiffness values between $1 \times 10^5$–$3 \times 10^5$ $N/m$, while buses and heavy trucks can reach sprung masses above 10,000 $kg$ with suspension stiffness in the order of $5 \times 10^5$–$1.5 \times 10^6$ $N/m$ [56, 57, 58]. In the design space illustrated in Figure 13a, these representative commercial vehicle configurations are indicated by white dots. As shown, they are concentrated in the low-stiffness region of the design space, with masses ranging from passenger car to heavy truck scales. In this region, the damage assessment performance is only moderate. This outcome is expected, since commercial vehicles are primarily tuned for ride comfort and safety rather than for maximising the transmissibility of bridge vibrations [59, 60, 58]. By contrast, the optimised purpose-built configuration (SV1) occupies a region associated with superior sensitivity to bridge damage, highlighting the advantage of tailoring vehicle dynamics specifically for inspection rather than relying on comfort-driven commercial designs.

It is worth noting that these design guidelines are intended for application in a semi-autonomous inspection vehicle developed by this research group (refer to Section 7). This vehicle is fully customisable, with all dynamic properties operating under controlled conditions, thereby rendering uncertainties associated with the vehicle properties negligible. Moreover, the inspection vehicle is equipped with high-quality, low-noise PCB-ICP393B05 sensors. As a result, sensor noise is not explicitly included in the model, given the negligible contribution of measurement noise from these sensors. In addition, the assumption of constant vehicle speed is justified, as the inspection vehicle is equipped with an electronic speed controller that ensures steady velocity across the bridge.

*5.3. Impact of Sensing Vehicle Characteristics in Drive-by Bridge Inspection*

In this section, the damage assessment performance of five different inspection vehicle configurations (e.g., SV1:SV5) within the design space presented in the previous section is compared. This chapter aims to demonstrate how inspection vehicle properties influence damage assessment performance. Table 6 lists the properties of the vehicles under comparison, which are also represented by red crosses in Figures 13a and 13b. The results of this study are presented in both dimensional (i.e., $k_v$ and $m_v$) and non-dimensional (i.e., $\mu$ and $\beta$) spaces. Vehicles SV1 and SV2 are selected to represent the optimal inspection vehicle configuration (see Section 5.2) and the vehicle used in previous studies [22, 23], respectively. In contrast, SV3, SV4, and SV5 are chosen to represent vehicles near the extrema of the design space. More specifically, SV3 represents a poorly performing vehicle (i.e., $\bar{W}_d \ll 0.01$) with a natural frequency close to the bridge's first natural frequency (i.e., $\beta \approx 1$). SV4 and SV5 are selected to represent vehicles with low mass and tyre-suspension stiffness and high mass and tyre-suspension stiffness, respectively.

Figure 14 presents examples of the pre-processed frequency spectra from the five inspection vehicles, SV1:SV5, for the two bridge conditions *HN* and *DM*. From the spectra of vehicles SV1 and SV2, a shift in the bridges natural frequency is identifiable (e.g., from 4.09 $Hz$ to 4.01 $Hz$) due to the damage. In contrast, for vehicle SV3, which has the poorest performance (reflected by the lowest $\bar{W}_d$), the frequency spectra for both the *HN* and *DM* conditions do not exhibit a perceptible variation in the bridge's natural frequency. This observation is attributed to SV3 operating near resonance with the first natural frequency of the structure (i.e., $\beta \approx 1$). A similar observation can be made from SV5, which operates close to resonance with the structure and has a large mass and tyre-suspension stiffness. From the $\overline{W}_d$ of SV4 (see Figure 13), it can be concluded that it performs better than SV3 and SV5. However, it is noticeable from the pre-processed spectrums that there is no apparent difference between the healthy and damaged samples, as in SV1 or SV2.

Table 6: Inspection vehicles properties.

| Inspection vehicle | $m_v$ (kg) | $k_v$ (N/m) | $\mu$ | $\beta$ |
|---|---|---|---|---|
| SV1 | 12,340 | 9.53 $\times 10^5$ | 0.0269 | 0.418 |
| SV2 | 16,200 | 4 $\times 10^5$ | 0.035 | 0.27 |
| SV3 | 2,344 | 54.86 $\times 10^5$ | 0.0051 | 1.0026 |
| SV4 | 995 | 2.45 $\times 10^5$ | 0.0022 | 0.2118 |
| SV5 | 19,151 | 76.1 $\times 10^5$ | 0.0417 | 1.097 |

Additionally, it is observed from Figure 14 that vehicles with larger masses and low tyre-suspension stiffness (i.e., SV1 and SV2) can more effectively distinguish between the *HN* and *DM* bridge conditions. This phenomenon occurs due to the larger mass inducing more significant vibration in the structure, which excites the first natural frequency of the bridge with a higher amplitude. These conditions are also observable in Figure 13. Furthermore, it can also be concluded that the vehicles that have a natural frequency similar to the first natural frequency of the bridge have a limited damage assessment performance regardless of their mass (e.g., SV3 and SV5).

After obtaining the pre-processed frequency spectra from inspection vehicles SV1:SV5, the unsupervised learning damage assessment model based on AAE is trained and tested for each vehicle. The corresponding damage assessment performance for each vehicle is presented in Figure 15. As previously discussed, inspection vehicles SV1 and SV2 can distinguish between the two bridge conditions, *HN* and *DM*, with only minor overlap between the healthy and damaged DI distributions. In contrast, vehicles SV3 and SV5 cannot detect the damaged state of the structure due to its proximity to resonance with the structures first natural frequency, presenting complete overlap between the *HN* and *DM* DI distributions.

Table 7 summarises the damage assessment metrics for each vehicle. Inspection vehicles SV3 and SV5 have the poorest damage assessment performance, compared to SV1 and SV2, which perform well in identifying damage by having $\overline{W}_d > 0.05$ and accuracy $> 90\%$. This is attributed to their significantly larger mass relative to SV3 and their operation away from resonance with the structure. The mean DI value for the *DM* condition is 3.8 times greater than that for the *HN* condition in the case of SV1, compared to 2.9 times for SV2. This indicates that SV1 has superior damage detection performance than SV2. Furthermore, SV1 exhibits the highest classification accuracy, corroborated by the $\overline{W}_d$ value, being 35% and 65% larger than those of SV2 and SV3, respectively. SV4 has an intermediate performance between the best-performing vehicle, SV1, and the worst-performing vehicle, SV3, reflected in a classification accuracy of 77.5%. However, from the performance of SV4, a conclusion on the bridge condition can not be stated, as significant overlap exists between the *HN* and *DM* samples. It is worth noting that the classification accuracy depends on the selected threshold, as stated in Section 4.3.1. It is also observed from Table 7, that the damage assessment performance is also reflected in the mean value of the DI from the *HN*. For the best-performing vehicle, SV1, the *HN* mean DI is approximately 21% lower than the mean DI for *HN* of the worst-performing vehicle, SV3.

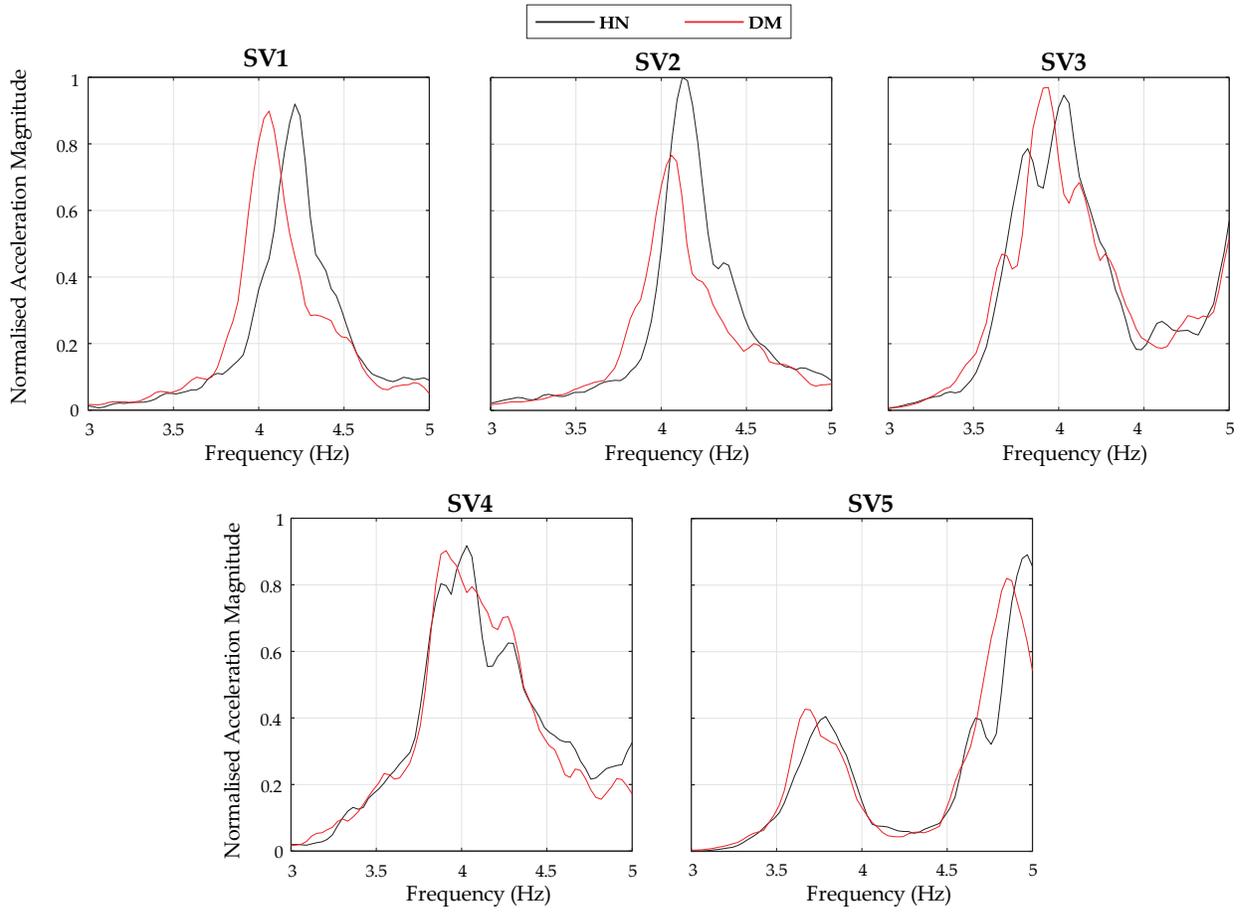

Figure 14: Comparison between healthy and damaged pre-processed frequency spectrums from the inspection vehicles SV1 to SV5.

Table 7: Damage assessment metrics summary for inspection vehicles SV1 to SV5.

| Vehicle | *HN* mean | *DM* mean | Threshold | Accuracy | $\overline{W}_d$ |
|---|---|---|---|---|---|
| SV1 | 0.0281 | 0.1063 | 0.040 | 95% | 0.0782 |
| SV2 | 0.0293 | 0.0852 | 0.042 | 94.5% | 0.0559 |
| SV3 | 0.0340 | 0.0367 | 0.046 | 52% | 0.0027 |
| SV4 | 0.0466 | 0.0720 | 0.065 | 77.5% | 0.0254 |
| SV5 | 0.0357 | 0.0443 | 0.051 | 53.5% | 0.0092 |

In conclusion, although the optimal vehicle identified in Section 5.2 has been shown to outperform other vehicles in the design space for damage identification, relying on a single sensing vehicle for a wide range of bridges is impractical. For that reason, it is suggested to use inspection vehicles with $\overline{W}_d > 0.04$ (see Figure 13). Ideally, it is suggested to use vehicles with $0.3 \leq \beta \leq 0.7$, considering that the lower $\beta$, the lower $\mu$.

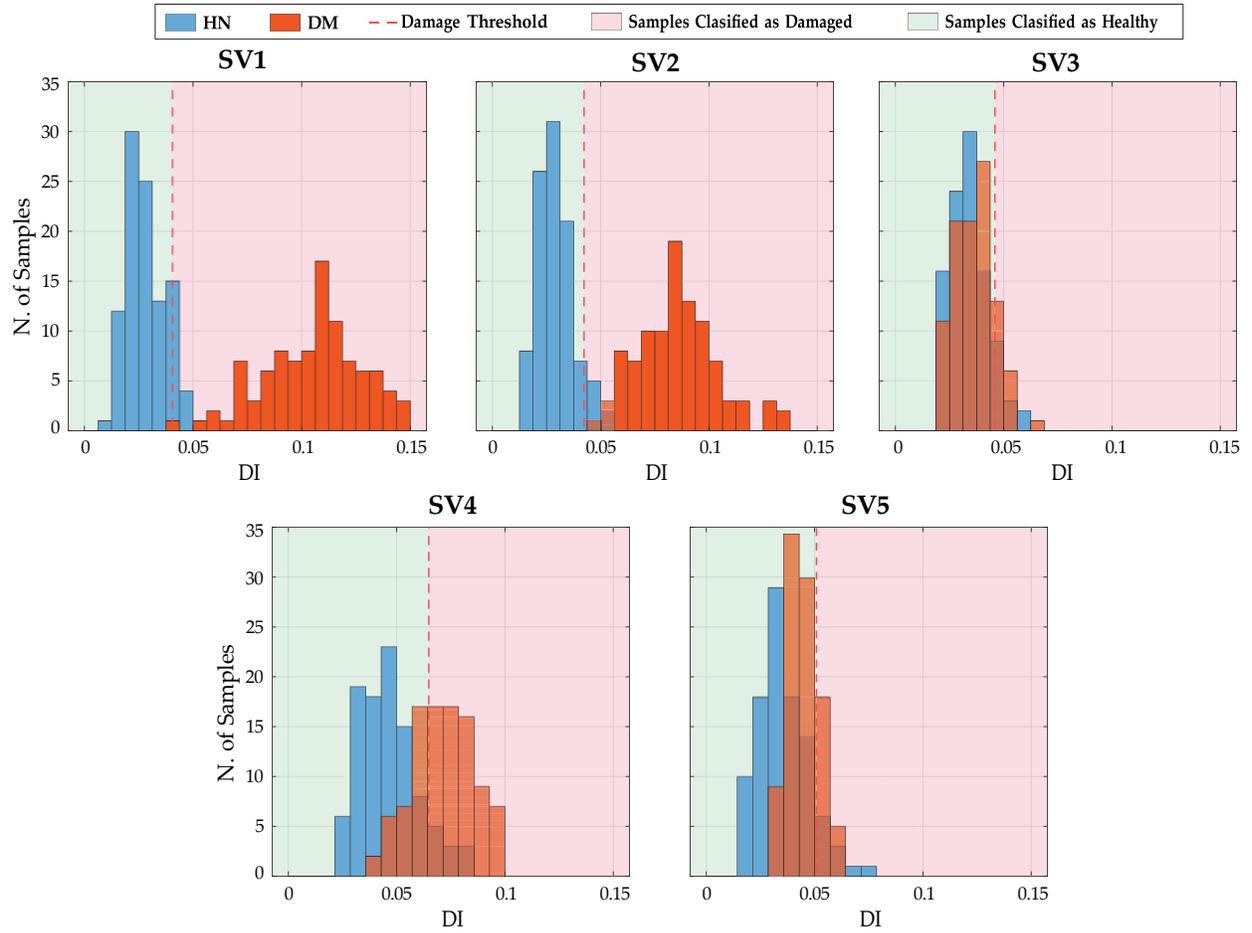

Figure 15: Comparison of damage assessment performance between inspection vehicle setups SV1 to SV5.

## 5.4. Assessment of Optimal Vehicle Configuration

### 5.4.1. Damage at Different Location

In previous sections, the damage assessment performance of various inspection vehicles was evaluated under a bridge with mid-span damage with $a_c = 10\%$ crack. In this section, the optimisation previously presented in Section 5.2 is assessed by evaluating the damage assessment performance of the five vehicles SV1:SV5, introduced in Section 5.3, including the optimal vehicle, SV1. To evaluate the performance, only the location of the damage on the bridge is varied, and the severity is kept (i.e., crack with $a_c = 10\%$ at $L/4$). Such damage produces a variation of 0.98% in the first natural frequency of the bridge. The classification results from the AAE framework are presented in Table 8, where it is appreciated that the damage assessment trend among the five inspection vehicles analysed is maintained as in Section 5.3.

Table 8: Damage assessment metrics summary for inspection vehicles SV1:SV5 with damage at $L/4$.

| Vehicle | Accuracy | $\overline{W}_d$ |
|---------|----------|------------------|
| SV1 | 80% | 0.016 |
| SV2 | 73% | 0.013 |
| SV3 | 50% | 0.005 |
| SV4 | 60.5% | 0.009 |
| SV5 | 51.5% | 0.003 |

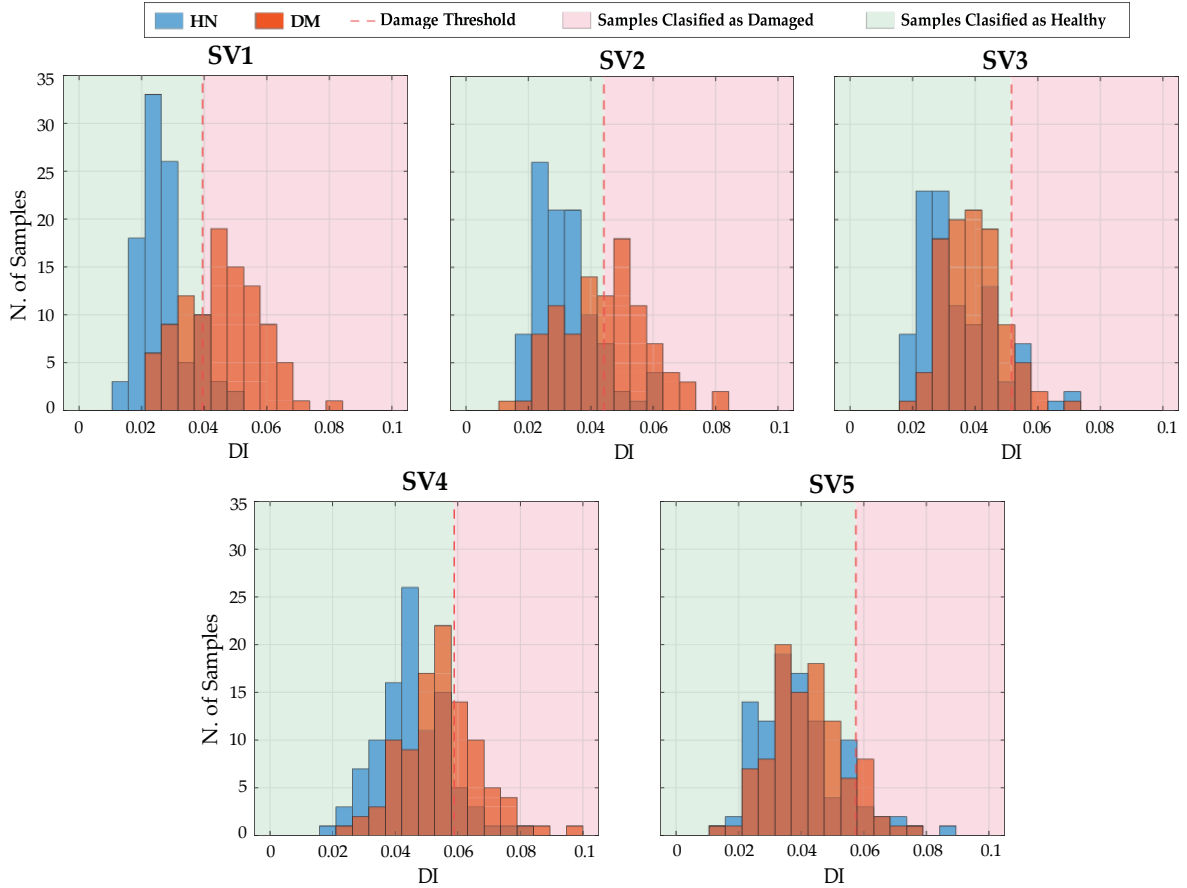

Figure 16: Comparison of damage assessment performance between inspection vehicle setups SV1:SV5 for a 10% beam height to crack depth ratio at one-quarter of the span.

Additionally, Figure 16 illustrates the damage assessment performance of SV1:SV5, as obtained using the framework detailed in Section 4. Due to the specific damage location and the minimal variation in the first natural frequency of the structure, there is an observable overlap between the *HN* and *DM* samples across all vehicles. This behaviour has also been observed and explained in [22], where SV2 displayed similar performance under the same damage assessment method. Nonetheless, the results are consistent with earlier findings presented in Section 5.3, reaffirming SV1 as the best-performing vehicle, with SV3 and SV5 exhibiting the weakest performance. Table 8 summarises the classification metrics for the five vehicles, highlighting that SV1 achieves the highest accuracy (i.e., 80%), further supporting previous conclusions. Additionally, SV1 records the highest $W_d$ value, indicating superior damage detection capabilities amongst the five vehicle configurations. On the other hand, SV4 shows an intermediate performance between the best and worst-performing vehicles (i.e., SV1 and SV2, respectively). However, a definitive assessment of the bridge's condition cannot be made due to the considerable overlap between the *HN* and *DM* samples. This consistency across metrics for different damage locations suggests that the optimisation procedure is robust and can reliably be extended to detect damage at various locations on the structure.

*5.4.2. Inspection Vehicle Validation on Different VBI System*

In this section, the inspection vehicle optimisation presented in Section 5.2 is assessed under a different target bridge. To this end, the bridge presented in [40, 61] (see Table 9) is inspected with two different vehicles, SV-N1 and SV-N2. The properties of SV-N1 and SV-N2, described in Table 10, were obtained based on the non-dimensional properties of vehicles SV1 and SV3 from the Kriging metamodel presented in Section 5.2 (i.e., the best and worst performing vehicles, respectively).

Table 9: Physical properties of the new bridge to be inspected.

| Property | Value | Property | Value |
|---|---|---|---|
| Span length, $L$ | 15 $m$ | Cross section area, $A$ | 7.5 $m^2$ |
| Young's modulus, $E$ | $3.5 \times 10^{10}$ $N/m^2$ | Second moment of inertia, $I_o$ | 0.527 $m^4$ |
| Mass per unit length, $\mu_b$ | 28, 125 $kg/m$ | Damping ratio, $\xi$ | 3% |
| First bridge's natural frequencies, $f_b$ | 5.65 $Hz$ | | |

Table 10: Inspection vehicle properties for new VBI system.

| Inspection Vehicle | $m_v$ (kg) | $k_v$ (N/m) | $\mu$ | $\beta$ |
|---|---|---|---|---|
| SV-N1 | 7,565 | 1.85 $\times 10^6$ | 0.0269 | 0.4180 |
| SV-N2 | 1,434 | 10.7 $\times 10^7$ | 0.0051 | 1.0026 |

To evaluate the damage assessment performance of the inspection vehicles SV-N1 and SV-N2, a damage level of $\alpha_c$ = 10% at $L/2$ is introduced to the bridge, which generates a variation of the first natural frequency of 2.3% (i.e., the natural frequency of the damaged bridge is 5.52 $Hz$). The damage assessment methodology described in Section 4 is then applied. Figure 17 illustrates the DI for both healthy and damaged samples recorded by SV-N1 and SV-N2. It is evident that inspection vehicle SV-N1, which shares similar non-dimensional properties ($\mu$ and $\beta$) with SV1 in Section 5.2, demonstrates superior damage assessment performance. This is reflected in the distinct distributions of the DI, with damaged samples exhibiting higher DI values and higher $W_d$ (see Table 11). In contrast, inspection vehicle SV-N2 is unable to effectively assess damage, as the DI values for healthy and damaged samples show substantial overlap. This limitation is further quantified by the lower $W_d$ value for SV-N2 compared to SV-N1, as presented in Table 11.

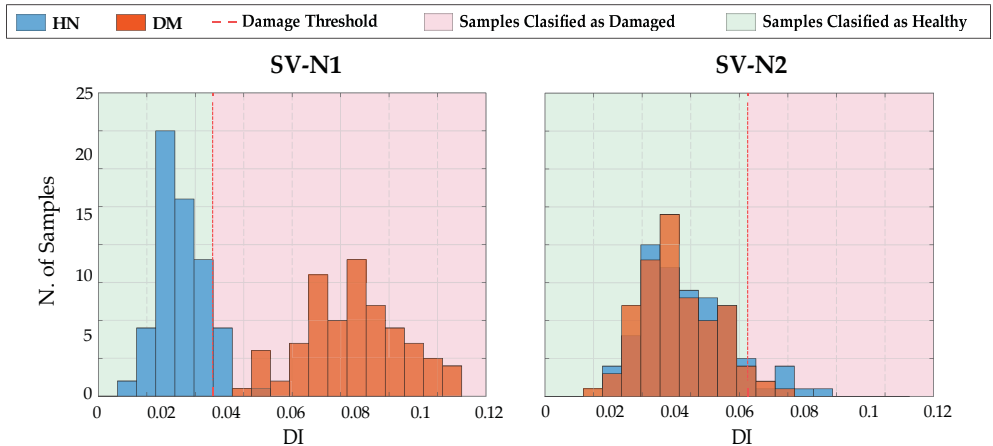

Figure 17: Comparison of damage assessment performance between inspection vehicle setups SV-N1, SV-N2 on a different VBI system.

Table 11: Damage assessment metrics summary for inspection vehicles SV-N1 and SV-N2.

| Inspection Vehicle | Accuracy | $W_d$ |
|---|---|---|
| SV-N1 | 95% | 0.0732 |
| SV-N2 | 47% | 0.0034 |

The superior performance of vehicle SV-N1 compared to SV-N2 is also evident from the better classification performance shown in Table 11, where SV-N1 is approximately 50% superior classification performance to SV-N2. These results confirm that the optimisation performed in Section 5.2 is robust and can be extended to various bridge configurations.

### 5.4.3. Experimental Validation

In this section, the inspection vehicle design guidelines presented in Section 5.2 are validated against the experimental results of Makki Alamdari et al. in [62]. In this work, the experimental tests performed in [62] are used to validate and enhance the rigour of the inspection vehicle design guidelines previously presented. In [62], the authors of the study suggest and verify a transmissibility-based approach for drive-by bridge inspection. This approach employs the dynamic response of a moving vehicle to identify changes in the modal frequencies of a bridge. They introduce a novel transmissibility index to quantify how effectively a vehicle captures bridge-induced vibrations and demonstrate, through both numerical simulations and laboratory experiments, that vehicles with superior transmissibility characteristics are more effective in identifying bridge vibration modes. The results confirm that, with appropriate vehicle dynamics and sufficient excitation, even small changes in bridge stiffness can be identified from vehicle response data.

To validate the experimental results presented in [62] (see Figure 18), VBI system is numerically modelled in this work. The two inspection vehicles used in the reference study are simulated and compared using the damage assessment approach developed in this work to demonstrate that the proposed design guidelines are consistent with the experimental observations. The properties of the experimental bridge are summarised in Table 12, and the dynamic characteristics of the two inspection vehicles (VM1 and VM2) are provided in Table 13.

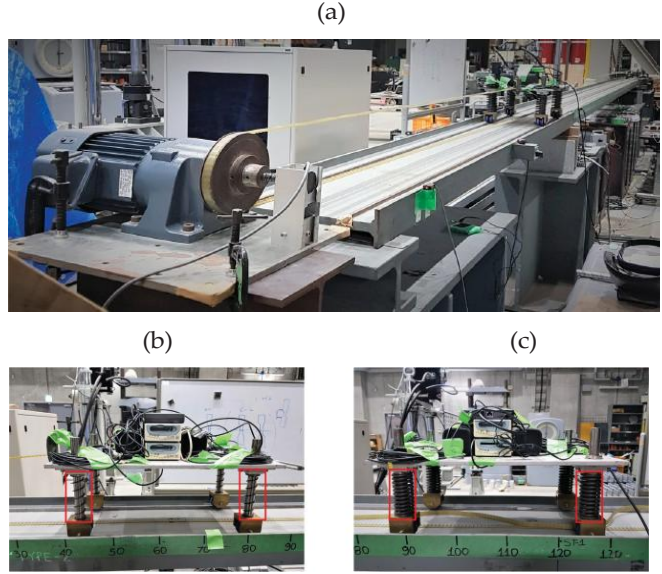

Figure 18: Experimental setup from the study in [62]: (a) the VBI system, (b) the VM1 inspection vehicle, and (c) the VM2 inspection vehicle.

Table 12: Physical properties of the experimental bridge model.

| Property | Value |
| --- | --- |
| Span length, $L$ | 5.4 m |
| Youngs modulus, $E$ | $2.1 \times 10^{11}$ N/m² |
| Density, $\rho_b$ | $7.8 \times 10^3$ kg/m³ |
| Cross-sectional area, $A$ | $7.04 \times 10^{-3}$ m² |
| Second moment of inertia, $I_o$ | $11.36 \times 10^{-7}$ m⁴ |
| First natural frequency, $f_b$ | 3.61 Hz |
| Damaged natural frequency, $f_{b_{damage}}$ | 3.66 Hz |

The framework devised in this work is used to numerically model the experimental setup described in [62], as previously mentioned. Figure 19 presents the DI for both healthy and damaged samples, recorded by vehicles VM1 and VM2. While damage is identifiable from the data collected by both vehicles, VM2 exhibits a clearer distinction between healthy and damaged states. This outcome may be attributed to the observation made in previous sections, where it was established that for a given $\mu$ (i.e., the vehicles in this setup have equal mass), the best-performing vehicles are those operating furthest from resonance conditions, that is when $\beta \approx 1$.

Table 13: Dynamic properties of the two inspection vehicles used in the experimental study.

| Inspection Vehicle | $m_v$ (kg) | $I_v$ (kgům$^2$) | $k_v$ (N/m) | $f_{v1}$ (Hz) | $\mu$ | $\beta$ |
|---|---|---|---|---|---|---|
| VM1 | 21.07 | 0.19 | 5,514 | 5.37 | 0.0712 | 1.49 |
| VM2 | 21.07 | 0.19 | 35,693 | 13.09 | 0.0712 | 3.63 |

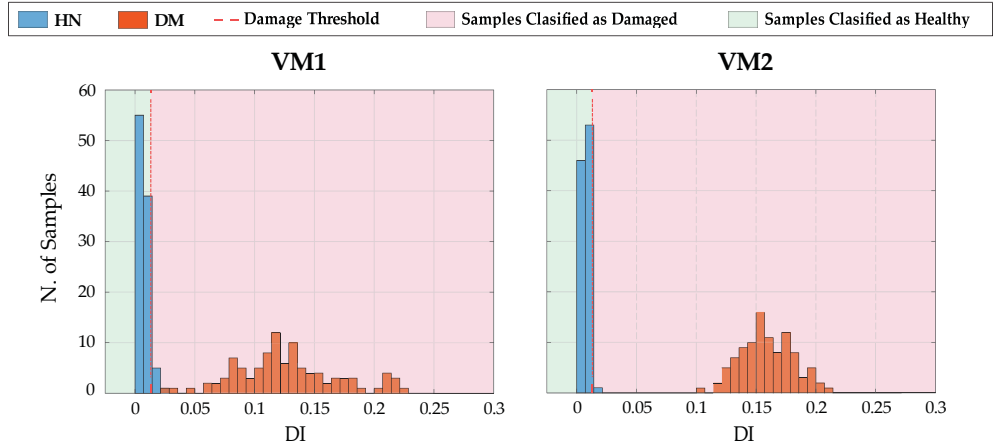

Figure 19: Comparison of damage assessment performance between inspection vehicle setups VM1 and VM2 from the experimental setup in [62].

Table 14: Damage assessment metrics summary for inspection vehicles VM1 and VM2.

| Inspection Vehicle | Accuracy | $W_d$ |
|---|---|---|
| VM1 | 95% | 0.1191 |
| VM2 | 95% | 0.1527 |

Furthermore, Table 14 presents the accuracy and $W_d$ values obtained for each vehicle. The results confirm that VM2 outperforms VM1, as indicated by its higher $W_d$ value. These findings are consistent with those reported in [62], where the authors demonstrated that VM2 exhibits more stable transmissibility, allowing for more accurate identification of interaction frequencies.

## 6. Validation on a Full-Scale Bridge

To validate the performance of the proposed methodology under nominal bridge condition, a benchmark study was conducted on the Bulli Colliery Bridge, a 23.9 $m$ steel girder pedestrian bridge located in New South Wales, Australia. The structure consists of four simply supported steel girders with a concrete deck, with an effective span of 22.5 $m$ and a width of 2.7 $m$. Figure 20 provides an overview of the bridge. The aim of this study is threefold: (1) to validate that the proposed custom-built inspection vehicle produces results consistent with direct sensing benchmarks, (2) to confirm that vehicle-related frequencies do not interfere with the frequency band containing the bridges fundamental vibration mode, and (3) to establish the nominal state of the structure through the application of the AAE methodology.

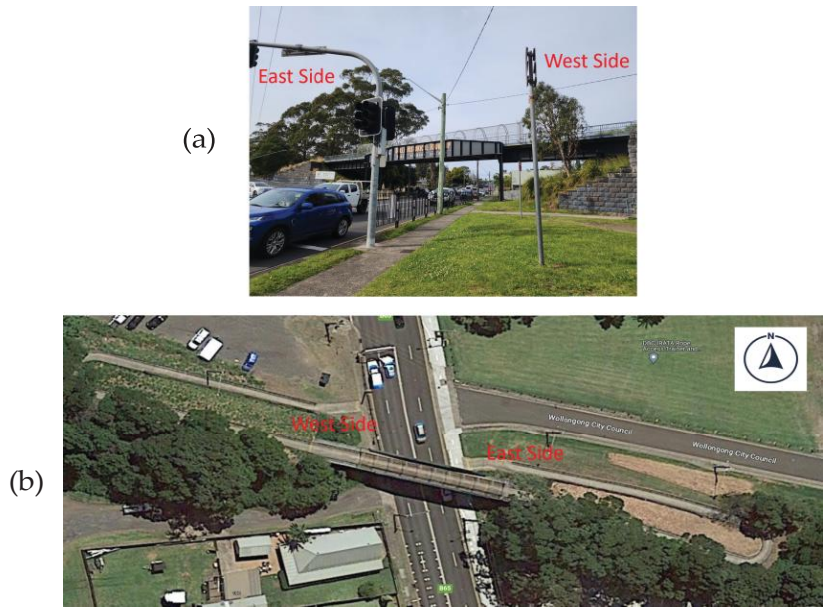

Figure 20: Bulli Colliery Bridge: (a) side view of the bridge facing south, (b) top view from Google Maps.

## 6.1. Direct Sensing Test

The nominal dynamic properties of the Bulli Colliery Bridge were first identified through direct sensing. A set of three wireless triaxial BeanAir accelerometers was deployed on the deck, mounted at quarter-span, mid-span, and three-quarter-span locations along the longitudinal axis. The sensors were fixed to the deck surface using adhesive mounts to ensure high-fidelity recordings. Data were collected during ambient excitation tests, lasting 1 minute with a sampling rate of 500 *Hz*. Figure 21a shows the experimental setup of the direct sensing configuration, while Figure 21b illustrates the drive-by inspection test, which will be discussed in the following subsection. Examples of the acceleration time series from the direct sensing tests are shown in Figure 22.

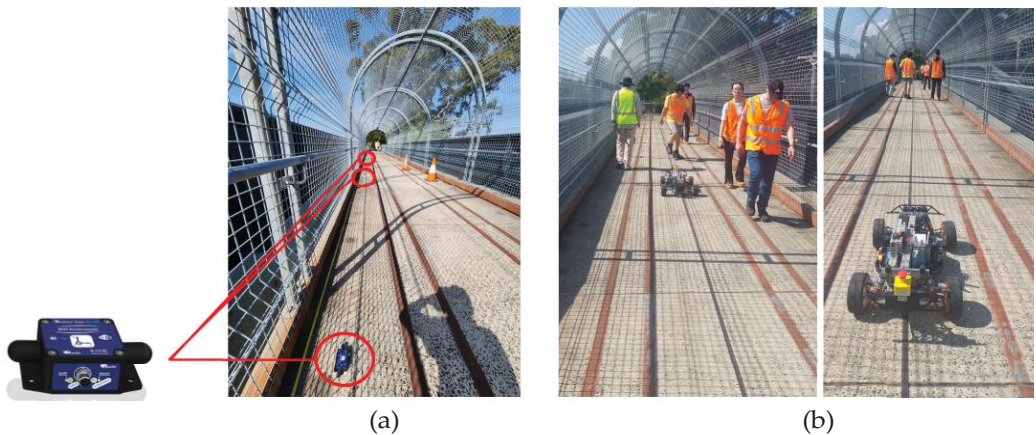

Figure 21: Bulli Colliery Bridge direct inspection tests: (a) direct test with three triaxial BeanAir sensors on the bridge, (b) drive-by inspection.

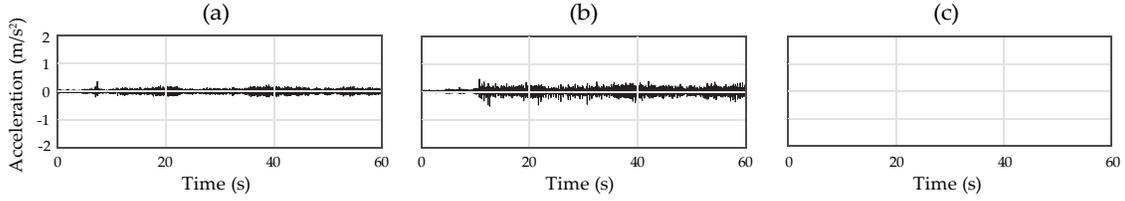

Figure 22: Acceleration time series from the sensors located at quarters of the span: (a) quarter, (b) mid-span, (c) three-quarters.

Then, the acceleration responses from the three sensors were further processed using singular value decomposition (SVD) to extract the dominant modal properties of the structure. Figure 23 presents the first singular value spectra obtained from five independent tests combining the sensor signals. The results consistently confirm the presence of a dominant mode at 6.7 *Hz*, which corresponds to the fundamental vibration frequency of the bridge under nominal condition. This frequency serves as a benchmark for evaluating the results of the indirect sensing tests presented in the following section.

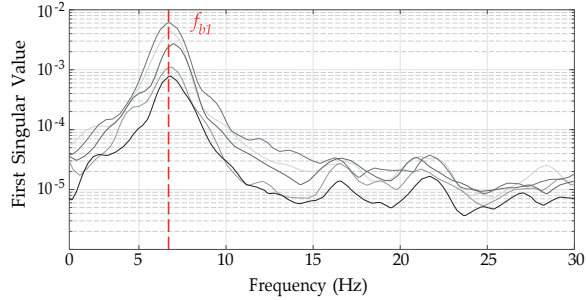

Figure 23: First singular value from five different direct inspection tests. The red dashed line represents the first natural frequency of the structure.

*6.2. Indirect Sensing with the Purpose-Built Vehicle*

To demonstrate that vehicle-related vibrations do not interfere with the frequency range of interest, which contains the bridges fundamental frequency, a series of validation runs were first performed on a smooth surface where no interaction with bridge exists. Figure 24a shows the acceleration response recorded from the front-right suspension arm during one such run. The corresponding SVD analysis considering the acceleration responses from the sensors on the four suspension arms, obtained from five repeated runs, is presented in Figure 24b. The results demonstrate that the dominant driving-related frequency component, $f_{vd} \approx 15$ *Hz*, lies outside the 010 *Hz* band that encompasses the bridges fundamental mode and that is later used in the assessment. This confirms that motor-induced vibrations from the sensing vehicle do not affect the frequency region relevant for assessing the structural condition.

Following this validation, indirect tests were carried out on the Bulli Colliery Bridge using the purpose-built sensing vehicle introduced in Section 7 (see Figure 21b). The vehicle, instrumented with accelerometers mounted on the suspension arms and operated at a constant speed of 0.17 *m/s*, crossed the bridge a total of 39 times, each producing acceleration responses sampled at 1,828 *Hz*. The collected responses were subsequently processed using

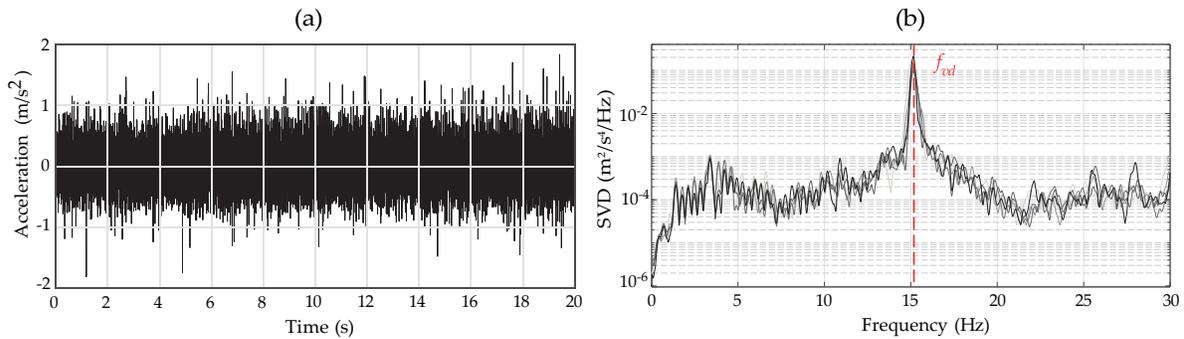

Figure 24: Illustration of: (a) acceleration time response from the front-right suspension arm, (b) first singular value result from five repeated runs combining all sensor responses. The red dashed line marks the identified driving frequency $f_{vd}$.

SVD, and the first singular values was filtered in the range of 0-10 *Hz* to capture the bridges fundamental vibration mode. The outcome of this analysis is illustrated in Figure 25, which presents the first singular value spectra obtained from five bridge crossings. The results consistently show a dominant peak at $f_{b1} \approx 6.7$ *Hz*, in excellent agreement with the frequency identified through direct sensing (see Figure 23). This confirms that the purpose-built sensing vehicle can reliably capture the bridges fundamental mode using indirect measurements.

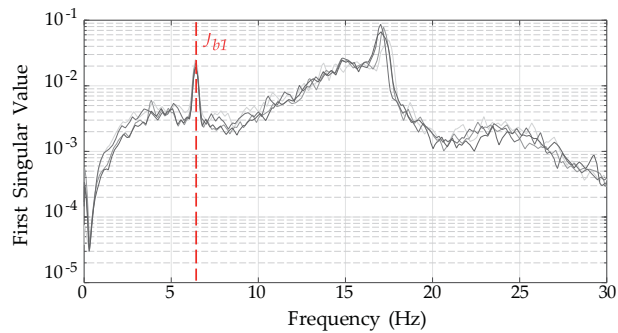

Figure 25: First singular values from five different indirect inspection tests on the Bulli Colliery Bridge. The red dashed lines represent the averaged first natural frequency of the structure $f_{b1}$.

The processed responses (i.e, the normalised first singular values from the responses collected by the sensing vehicle) were subsequently analysed using the AAE methodology. Training and testing samples were selected randomly using a 80:20 ratio, respectively. The AAE was able to reconstruct the spectral features of the nominal bridge state with high fidelity, demonstrating the models capability to reproduce the nominal frequency domain representations of the signals. Figure 26 shows an example of a reconstructed normalised first singular value from the Bulli Bridge indirect tests. The close alignment between the original and reconstructed signals provides strong evidence that the methodology captures the nominal structural response.

To quantify the performance across all the vehicle crossings, the reconstruction error distributions from the test set is shown in Figure 27. The results indicate that all test samples fall within the nominal classification range, thereby these findings demonstrate that the purpose-built vehicle enables reliable identification of the bridges nominal state.

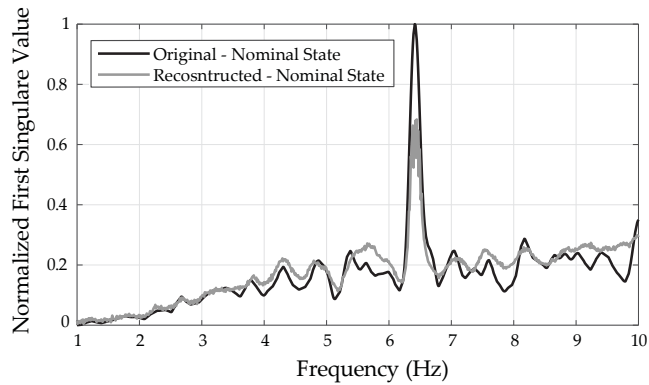

Figure 26: Reconstruction of a frequency spectrum sample using the AAE.

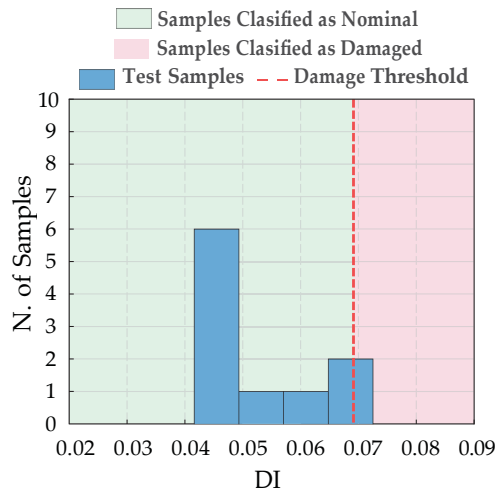

Figure 27: Reconstruction errors from the test set.

This benchmark study on the Bulli Colliery Bridge confirmed three main outcomes: (1) the custom-built sensing vehicle produces results consistent with direct measurements, (2) vehicle-induced frequencies do not interfere with the frequency band containing the bridges fundamental mode, and (3) the nominal state of the structure can be reliably verified using the AAE. These findings demonstrate the feasibility of the proposed methodology for indirect bridge assessment.

Nonetheless, some practical limitations remain for full-scale implementation of this sensing vehicle concept, including constraints on operating speed, the limited number of recordings used in this study, and the need for further investigations under varying vehicle properties. Future tests on additional structures, with varied vehicle speeds and configurations, will be carried out to further consolidate and extend the sensing vehicle optimisation results (see Section 7).

## 7. Future Work

The design guidelines presented in this study are intended for use in future drive-by bridge inspection applications, utilising a newly developed semi-autonomous sensing vehicle designed by this research group. The authors acknowledge the limitations of applying specific design guidelines to conventional vehicles. However, as previously noted, these guidelines are intended for a customisable vehicle, in which all dynamic properties can be fully controlled and adjusted as required. Figure 28 depicts the sensing vehicle designated for the forthcoming experimental work, equipped with four high-quality, low-noise PCB-ICP393B05 sensors and a speed controller that maintains a uniform velocity while traversing the bridge.

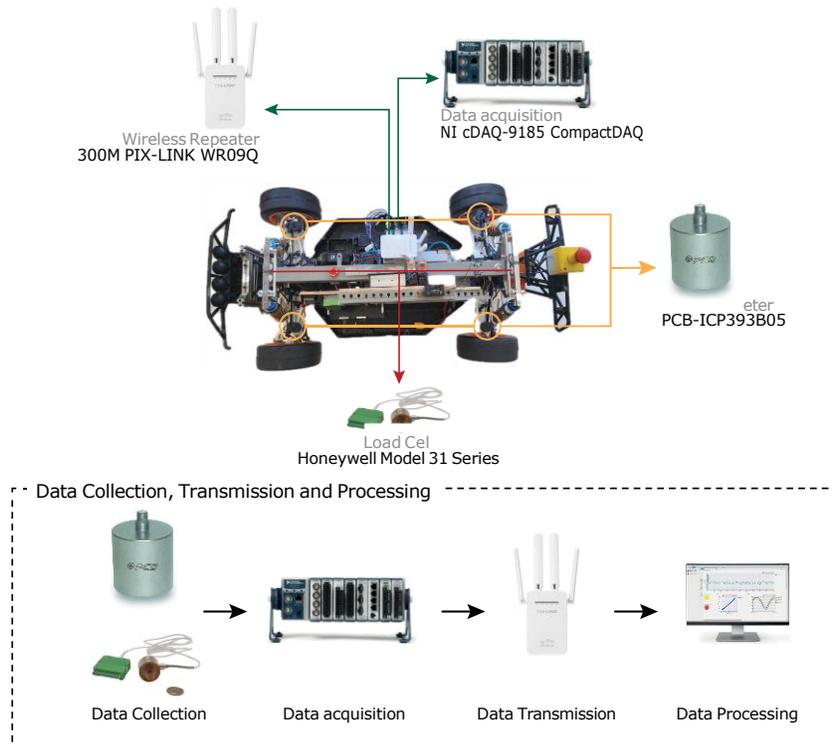

Figure 28: Inspection vehicle setup for drive-by bridge inspection.

Furthermore, this study aligns with the prevailing trend in the drive-by bridge inspection research, where environmental conditions are often overlooked, and the sensing speed is typically low [8]. Future work should therefore consider incorporating additional variables such as temperature fluctuations, wind effects, and variations in sensing vehicle speed. It is also worth noting that the proposed sensing methodology may require temporary partial road closures while the sensing vehicle crosses the bridge; consequently, the inclusion of traffic effects is beyond the scope of this study.

## 8. Conclusion

This research paper aimed to establish a framework for optimising inspection vehicles used in drive-by bridge inspections. A novel unsupervised damage assessment methodol-ogy employing adversarial autoencoders (AAE) was implemented, leveraging the frequency spectrum of the contact point response to identify damage on the bridge. The research indicated that specific vehicle layouts excel in damage detection compared to others. The performance was measured using the Wasserstein distance ($W_d$) between the distributions of reconstructed healthy and damaged samples by the AAE. A Kriging meta-model was subsequently developed to estimate $W_d$, enabling the determination of the ideal vehicle mass and stiffness that maximise $W_d$ in both dimensional and non-dimensional spaces, hence enabling the results to be applicable to various bridge configurations. The investigation indicated that inspection vehicles resonating with the bridge's initial natural frequency exhibit the lowest capacity for damage detection.

In contrast, vehicles exhibiting vehicle-bridge frequency ratios ranging from 0.3 to 0.7 were the most efficacious for damage detection. Furthermore, it was noted that vehicles with reduced mass demand proportionately lower natural frequencies for optimal performance. The study further indicated that the suggested optimisation approach may be adapted to identify damages at various locations on the bridge and across different VBI systems, hence enhancing the framework's robustness and applicability to real-world situations. It is crucial to recognise that the ideal vehicle configurations are significantly influenced by the damage assessment methodology employed, and different approaches may produce distinct optimal inspection vehicles. The favourable results of this study offer opportunities for more investigation and improvement of the framework. The authors highlight multiple avenues for future research. The optimisation framework will be augmented to incorporate supplementary vehicle characteristics, including vehicle damping, axle distance, and the position of the centre of mass. The authors intend to evaluate the optimisation framework on several large-scale bridges in New South Wales (NSW), Australia, utilising a specially designed inspection vehicle.

## 9. Acknowledgment

The authors express gratitude to the Australian Research Council (ARC) for their assistance through the Discovery Early Career Researcher Award (DECRA) initiative, grant number DE210101625. This research received financing from the Australian Research Council's Industrial Transformation Research Programme (IH210100048).

**Appendix A. VBI Formulation**

The bridge is modelled as a simply supported Euler-Bernoulli beam, incorporating a stochastic road profile denoted by $r(x_{v_i})$, where $x_{v_i}$ corresponds to the location of the $i$-th axle along the bridge span.

The equations governing the dynamics of the two-degree-of-freedom vehicle system are provided in Equations A.1 and A.2:

$$\begin{aligned}
m_v \ddot{z}_v(t) &+ c_{v_1} [\dot{z}_v(t) + d_1 \dot{\theta}_v(t) - \{\iota_b(x_{v_1})\}^T \{\dot{Z}_b(t)\} + v r'(x_{v_1})] \\
&+ c_{v_2} [\dot{z}_v(t) - d_2 \dot{\theta}_v(t) - \{\iota_b(x_{v_2})\}^T \{\dot{Z}_b(t)\} + v r'(x_{v_2})] \\
&+ k_{v_1} [z_v(t) + d_1 \theta_v(t) - \{\iota_b(x_{v_1})\}^T \{Z_b(t)\} + r(x_{v_1})] \\
&+ k_{v_2} [z_v(t) - d_2 \theta_v(t) - \{\iota_b(x_{v_2})\}^T \{Z_b(t)\} + r(x_{v_2})] = 0
\end{aligned} \quad (A.1)$$

$$\begin{aligned}
I_v \ddot{\theta}_v(t) + d_1 \Big\{ c_{v_1} \big[ \dot{z}_v(t) + d_1 \dot{\theta}_v(t) - \{\iota_b(x_{v_1})\}^T \{\dot{Z}_b(t)\} + v r'(x_{v_1}) \big] \\
+ k_{v_1} \big[ z_v(t) + d_1 \theta_v(t) - \{\iota_b(x_{v_1})\}^T \{Z_b(t)\} + r(x_{v_1}) \big] \Big\} \\
- d_2 \Big\{ c_{v_2} \big[ \dot{z}_v(t) - d_2 \dot{\theta}_v(t) - \{\iota_b(x_{v_2})\}^T \{\dot{Z}_b(t)\} + v r'(x_{v_2}) \big] \\
+ k_v^2 \big[ z_v(t) - d_2 \theta_v(t) - \{\iota_b(x_{v}^2)\}^T \{Z_b(t)\} + r(x_v{}^2) \big] \Big\} = 0
\end{aligned} \quad (A.2)$$

Let $\{Z_b(t)\} = \{Z_{b_1}(t), Z_{b_2}(t), \ldots, Z_{b_n}(t)\}$ represent the column vector of nodal displacements of the bridge, where $n$ is the total number of bridge degrees of freedom. The vector $\{\iota_b(x_v)\}$ contains polynomial interpolation functions that define the displacement of the bridge at the axle contact point $x_{v_i}$. Here, the overdot indicates time differentiation, while the prime denotes spatial differentiation.

The dynamic equilibrium equations for the bridge are formulated using the finite element method, incorporating the mass, damping, and stiffness matrices $[\mathbf{M}_b]_{n \times n}$, $[\mathbf{C}_b]_{n \times n}$, and $[\mathbf{K}_b]_{n \times n}$, respectively, as shown in Equation A.3:

$$\begin{aligned}
[\mathbf{M}_b]\{\ddot{Z}_b(t)\} + [\mathbf{C}_b]\{\dot{Z}_b(t)\} + [\mathbf{K}_b]\{Z_b(t)\} \\
+ \{\iota_b(x_{v_1})\}R_1(t) + \{\iota_b(x_{v_2})\}R_2(t) = 0
\end{aligned} \quad (A.3)$$

The contact forces at the axle positions $x_{v_1}$ and $x_{v_2}$ are defined by $R_1(t)$ and $R_2(t)$ as follows:

$$\begin{aligned}
R_1(t) = -c_{v_1} \big[ \dot{z}_v(t) - \{\iota_b(x_{v_1})\}^T \{\dot{Z}_b(t)\} + v r'(x_{v_1}) \big] \\
- k_{v_1} \big[ z_v(t) - \{\iota_b(x_{v_1})\}^T \{Z_b(t)\} + r(x_{v_1}) \big] + \left(\frac{d_2}{d}\right) m_v g
\end{aligned} \quad (A.4)$$

$$\begin{aligned}
R_2(t) = -c_{v_2} \big[ \dot{z}_v(t) - \{\iota_b(x_{v_2})\}^T \{\dot{Z}_b(t)\} + v r'(x_{v_2}) \big] \\
- k_{v_2} \big[ z_v(t) - \{\iota_b(x_{v_2})\}^T \{Z_b(t)\} + r(x_{v_1}) \big] + \left(\frac{d_1}{d}\right) m_v g
\end{aligned} \quad (A.5)$$

Here, $g$ denotes gravitational acceleration. The interaction between the vehicle and the bridge is described by coupling Equations A.1, A.2, and A.3, as discussed in [63]. The resulting coupled system of differential equations can be solved iteratively using established time integration schemes, such as the Newmark-Beta method [64].